\documentclass[11pt]{article}

\usepackage{microtype} 
\usepackage{booktabs}  
\usepackage{url}  

\usepackage{amsmath}
\usepackage{amsthm}

\usepackage{graphicx}
\usepackage[inline]{enumitem}
\usepackage{makecell}
\usepackage[linesnumbered,ruled,vlined]{algorithm2e}
\usepackage{float}

\usepackage{xcolor}
\definecolor{changedcolor}{rgb}{0.42,0.27,0.57}

\SetNlSty{textbf}{\color{black}}{}

\SetCommentSty{mycommentfont}

\usepackage{amsmath}
\usepackage{mathtools}  
\usepackage{amsfonts}

\newcommand{\nnunet}{nnU-Net}
\newcommand{\conv}{nnU-Net (Conv)}
\newcommand{\resm}{nnU-Net (ResM)}
\newcommand{\resl}{nnU-Net (ResL)}
\newcommand{\medsam}{MedSAM}
\newcommand{\medsamtwo}{MedSAM2}
\newcommand{\autonnunet}{Auto-nnU-Net}

\newcommand{\hpo}{\textsc{HPO}}
\newcommand{\hponas}{\textsc{HPO+NAS}}
\newcommand{\hpohnas}{\textsc{HPO+HNAS}}

\newcommand{\odsc}{\mbox{\textit{1 - DSC}}}

\newcommand{\Dds}[1]{%
    \ifcsname D-#1\endcsname
        \csname D-#1\endcsname
    \else
        \textbf{Unknown Dataset:} #1
    \fi
}

\expandafter\def\csname D-01\endcsname{D01}
\expandafter\def\csname D-02\endcsname{D02}
\expandafter\def\csname D-03\endcsname{D03}
\expandafter\def\csname D-04\endcsname{D04}
\expandafter\def\csname D-05\endcsname{D05}
\expandafter\def\csname D-06\endcsname{D06}
\expandafter\def\csname D-07\endcsname{D07}
\expandafter\def\csname D-08\endcsname{D08}
\expandafter\def\csname D-09\endcsname{D09}
\expandafter\def\csname D-10\endcsname{D10}

\usepackage[final]{automl}

\usepackage[%
  backend=biber,
  natbib=true,
  style=authoryear-comp,
  citestyle=authoryear, 
  bibstyle=authoryear,  
  dashed=false,  
  date=year,    
  urldate=ymd,   
  sorting=nyt,   
  sortcites=false,
  natbib=true,
  maxcitenames=1,  
  giveninits=true,
  maxcitenames=1,
  maxbibnames=5,
  uniquename=init,
  uniquelist=false,     
  doi=false,
  url=true,
  isbn=false,
]{biblatex}
\DeclareNameAlias{sortname}{given-family}

\bibliography{strings,shortproc,global,bib-refs} 

\title{\autonnunet: Towards Automated Medical Image Segmentation}

\author[1,5]{\nameemail{Jannis Becktepe\textsuperscript{\dag}}{jannis.becktepe@tu-dortmund.de}}
\author[1]{\nameemail{Leona Hennig}{l.hennig@ai.uni-hannover.de}}
\author[2,4]{\nameemail{Steffen Oeltze-Jafra}{Oeltze-Jafra.Steffen@mh-hannover.de}}
\author[1,3,4]{\nameemail{Marius Lindauer}{m.lindauer@ai.uni-hannover.de}}

\affil[1]{Institute of AI, Leibniz University Hannover}
\affil[2]{Peter L. Reichertz Institute for Medical Informatics, Hannover Medical School}
\affil[3]{L3S Research Center}
\affil[4]{CAIMed: Lower Saxony Center for AI \& Causal Methods in Medicine}
\affil[5]{Lamarr Institute for Machine Learning and Artificial Intelligence}

\hypersetup{
  pdfauthor={},
  pdftitle={},
  pdfsubject={},
  pdfkeywords={}
}

\begin{document}

\begingroup
\renewcommand\thefootnote{\dag}
\footnotetext{Work was conducted at Institute of AI, Leibniz University Hannover.}
\endgroup

\maketitle

\begin{abstract}
Medical Image Segmentation (MIS) includes diverse tasks, from bone to organ segmentation, each with its own challenges in finding the best segmentation model. 
The state-of-the-art AutoML-related MIS-framework nnU-Net automates many aspects of model configuration but remains constrained by fixed hyperparameters and heuristic design choices.
As a full-AutoML framework for MIS, we propose \autonnunet{}, a novel \nnunet{} variant enabling hyperparameter optimization~(HPO), neural architecture search~(NAS), and hierarchical NAS~(HNAS).
Additionally, we propose Regularized PriorBand to balance model accuracy with the computational resources required for training, addressing the resource constraints often faced in real-world medical settings that limit the feasibility of extensive training procedures.
We evaluate our approach across diverse MIS datasets from the well-established Medical Segmentation Decathlon, analyzing the impact of AutoML techniques on segmentation performance, computational efficiency, and model design choices.
The results demonstrate that our AutoML approach substantially improves the segmentation performance of \nnunet{} on 6 out of 10 datasets and is on par on the other datasets while maintaining practical resource requirements.
Our code is available at \url{https://github.com/automl/AutoNNUnet}.
\end{abstract}

\section{Introduction}
\label{sec:intro}

Machine learning (ML) plays a key role in modern healthcare, enabling accurate diagnoses~\citep{de_fauw-nature18,bernard-ieeeMedImg18,khan-expBioMed23, wang-nature24}, early cancer detection~\citep{cao-nature23}, and scientific discovery~\citep{falk-nature19}.
Medical image segmentation (MIS) aims to identify anatomical structures in medical scans but is challenging due to datasets variations, class imbalances, and task-specific constraints~\citep{litjens-mia17,isensee-nature19,ali-medImg24}.

Self-configuring methods reduce the need for manual tuning by adapting models for a given dataset~\citep{ali-medImg24}.
\nnunet{}~\citep{isensee-nature19} has emerged as a state-of-the-art framework that automatically configures U-Net-based architectures to achieve strong segmentation performance.
However, \nnunet{} surprisingly relies on some fixed hyperparameters and manually designed heuristics, which limit flexibility and may not always yield optimal results across datasets~\citep{bergstra-jmlr12a,quinton-springer24}.

In this work, we leverage automated machine learning~(AutoML)~\citep{hutter-book19a} to address these challenges and perform a large-scale study on the impact of AutoML on MIS.
We introduce \autonnunet{}, a novel variant of \nnunet{} that integrates AutoML to enable hyperparameter optimization~(HPO) and neural architecture search~(NAS) for \nnunet{}.
By combining PriorBand~\citep{mallik-neurips23a} with multi-objective optimization~\citep{karl-evolearn23a}, we introduce Regularized PriorBand for Joint Architecture and Hyperparameter Search (JAHS)~\citep{awad-arxiv23}, addressing the growing interest in resource efficiency in MIS as highlighted by recent work~\citep{rayed-infmed24}.
Our study evaluates \autonnunet{} on the Medical Segmentation Decathlon (MSD) datasets~\citep{simpson-arxiv19,antonelli-nature22}, providing insights into the impact of optimization strategies, hyperparameter importance, and dataset characteristics.
Notably, unlike most studies on AutoML for MIS~\citep{ali-medImg24}, we report results for all ten MSD datasets, providing a more thorough assessment of generalizability and robustness across diverse medical segmentation challenges.

In this work, we make the following contributions:
\begin{enumerate}
    \item \textbf{\autonnunet{} for AutoML-driven MIS.} We propose a novel framework that automates key design decisions in \nnunet{} for flexible and structured HPO and NAS.
    \item \textbf{Efficient optimization with Regularized PriorBand.} We introduce Regularized PriorBand, which incorporates training runtime as an optimization objective to reflect real-world constraints, where limited resources and frequent retraining make efficient training essential.
    It selects slower models only if they improve accuracy, and inherently yields trade-off solutions.
    \item \textbf{Extensive evaluation across all ten MSD datasets.} We analyze the impact of AutoML on segmentation accuracy, including hyperparameter importance and dataset transferability, enabling a deeper understanding of generalization behavior and guiding the design of more robust, efficient models across diverse medical imaging tasks.
\end{enumerate}

\section{Background on Image Segmentation}
\label{sec:background}

Following \citet{szeliski-book22}, \textit{semantic segmentation} refers to partitioning an image into regions associated with specific classes.
We use the term \textit{image segmentation} interchangeably to describe this task, where each pixel is labeled to enable structured analysis of visual data.

\textit{Medical image segmentation} (MIS) involves partitioning medical images, e.g., magnetic resonance imaging (MRI) or computer tomography (CT) scans, to identify areas of interest, incl. organs or potentially malicious structures such as tumors~\citep{antonelli-nature22}.
In practice, automated segmentation assists clinicians by accurately identifying critical areas for patient treatment~\citep{liang2019deep}.
Recent MIS datasets focus on foreground classes, treating the background as a single, excluded class~\citep{menze-ieee15,heller-arxiv19,simpson-arxiv19,antonelli-nature22}.
Unlike natural image segmentation, MIS faces challenges like limited availability of training data, class imbalances, small or branching anatomical structures, weak boundaries, and variable intensity distributions. 
The segmentation of 3D images consisting of \textit{voxels} (volume pixels) from MRI and CT scans further increases the segmentation complexity and computational demands of MIS~\citep{isensee-nature19,ali-medImg24}.

\section{Related Work}
\label{sec:related}

In this section, we review previous work on core components of \autonnunet{}: self-configuring segmentation frameworks, HPO, NAS, and multi-objective optimization for MIS. Our work aims to unify them into a comprehensive AutoML framework tailored to the task at hand.

\paragraph{\nnunet{}}
Self-configuring frameworks address the challenge of designing and tuning MIS models for a given task and dataset.
Building on U-Net's success~\citep{ronneberger-miccai15a}, \citet{isensee-nature19} introduced \nnunet{}, which optimizes an U-Net for a given task. Like CASH~\citep{thornton-kdd13a}, it jointly selects training hyperparameters and the final model or ensemble for inference.
\nnunet{} achieves this by leveraging dataset \textit{meta-features} common in AutoML~\citep{vanschoren-automlbook19a}.

We focus on the self-configuration mechanism of \nnunet{}, omitting pre- and post-processing steps.
The pipeline consists of three phases:
\begin{enumerate*}[label=(\roman*)]
    \item \textbf{Experiment planning}, where rule-based hyperparameter selection leverages dataset properties;
    \item \textbf{Training}, where 2D U-Net, 3D U-Net, and, if needed, a 3D U-Net cascade are trained using 5-fold cross-validation;
    \item \textbf{Inference}, where the best-performing model or ensemble is selected based on validation scores.
\end{enumerate*}
\nnunet{} relies on three types of hyperparameters: 
\begin{enumerate*}[label=(\roman*)]
    \item \textbf{Fixed} (e.g., learning rate, optimizer, loss function); 
    \item \textbf{Rule-baesd} (e.g., preprocessing, network topology);
    \item \textbf{Heuristic} (e.g., ensemble selection, post-processing).
\end{enumerate*}

\paragraph{Hyperparameter Optimization (HPO) for MIS}
In general, there is little work on HPO for MIS.
\citet{yang-miccai19} propose reinforcement learning for optimizing the data augmentation and learning rate of a 3D segmentation model.
\citet{quinton-springer24} apply HPO to different models, including \nnunet{}, by subsequently performing Bayesian optimization~(BO) for three groups of hyperparameters:
\begin{enumerate*}[label=(\roman*)]
    \item patch size,
    \item data pre-processing and augmentation, and
    \item loss function and optimizer.
\end{enumerate*}

\paragraph{Neural Architecture Search (NAS) for MIS}
Various NAS methods have been adapted for MIS.
Several approaches build on DARTS~\citep{liu-iclr19a} and apply differentiable NAS to encoder-decoder-based MIS models~\citep{weng-iee19,zhu-iee19,he-cvpr21}.
Another approach adopts a coarse-to-fine strategy for U-Net-shaped networks, first optimizing the overall topology before refining cell-level operations~\citep{yu-cvpr20}.
Evolutionary and graph-based NAS methods have also been proposed for MIS, using genetic algorithms~\citep{hassanzadeh-sac20,khouy_jpm2023,yu-cbm23} and graph representations of architectures that are optimized or expanded during training to reduce search time and improve flexibility~\citep{liu-cc23,qin-cmig23}.

\paragraph{Multi-Objective Optimization and Joint HPO and NAS for MIS}
Prior work has applied multi-objective NAS to MIS to balance performance and resource constraints~\citep{calisto-neuro20,lu-miccai22}, but without tuning hyperparameters. \citet{yang-iccv21} combine HPO and NAS via surrogate models to optimize U-Net configurations, but do not consider resource efficiency.

\section{\autonnunet{} for MIS}
\label{sec:approach}

In this work, we present \autonnunet{}, a novel approach that integrates AutoML methods into \nnunet{}.
Furthermore, we introduce Regularized PriorBand to enable efficient Joint Architecture
 and Hyperparameter Search (JAHS)~\citep{awad-arxiv23} in \autonnunet{}.

\subsection{Integrating AutoML Methods into \nnunet{}}
\label{sec:approach:autonnunet}

\begin{figure}[tb]
	\includegraphics[alt={Block diagram of the Auto-nnU-Net framework. It shows the flow of information starting from a dataset fingerprint and input configuration, consisting of hyperparameter configuration (lambda) and architecture (A). These inputs feed into the AutoExperimentPlanner and CFGUNet modules within Auto-nnU-Net. The pipeline then proceeds to generate nnU-Net plans, followed by training variants of U-Net (2D, 3D, and 3D cascade) using AutoNNUNetTrainer. The resulting models are evaluated against objectives (e.g., accuracy and runtime). The performance feedback loop informs (Regularized) PriorBand, which iteratively updates the hyperparameter and architecture configurations.}, width=\textwidth]{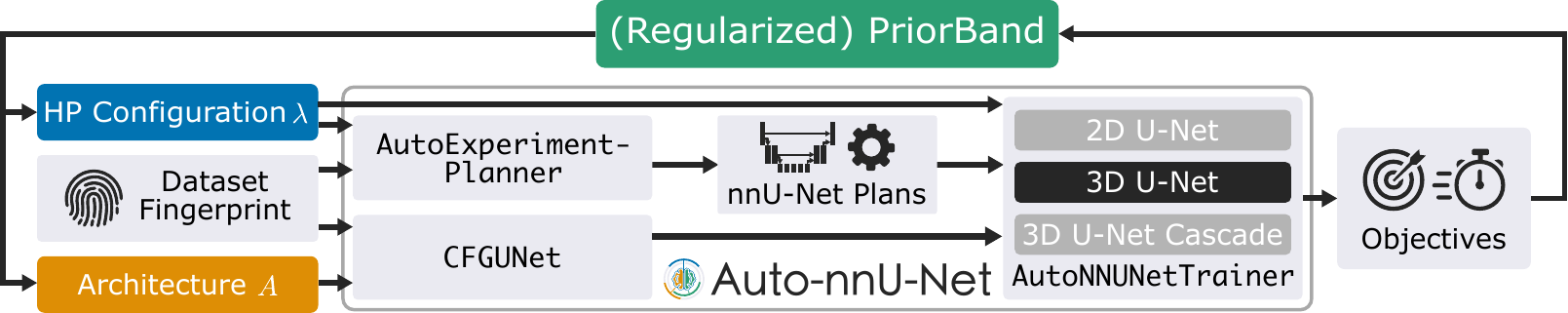}
	\caption{Overview of the \autonnunet{} framework: Given a hyperparameter configuration $\lambda$, architecture $A$, and dataset fingerprint, the \texttt{AutoExperimentPlanner} and \texttt{CFGUNet} generate nnU-Net training plans and model architecture, respectively. The \texttt{AutoNNUNetTrainer} then trains the selected model, providing runtime and validation score as objectives to the PriorBand optimizer. For details, see Appendix~\ref{sec:appendix:approach:autonnunet}. For more details, see Appendix~\ref{sec:appendix:approach:autonnunet}.}
	\label{fig:related:autonnunet_overview}
\end{figure}

\nnunet{} provides robust segmentation pipelines, including data pre-processing, experiment planning, training, and inference.
However, its fixed and rule-based hyperparameters limit configurability.
To address these limitations, we propose \textbf{\autonnunet{}}, which enhances \nnunet{} with flexible experiment planning and training.
Figure~\ref{fig:related:autonnunet_overview} shows an overview of our framework.
Unlike nnU-Net, \autonnunet{} takes hyperparameter and architecture configurations as inputs, enabling JAHS.
It returns both generalization error and training runtime to allow the optimization process to account for both segmentation performance and computational efficiency.

\subsection{Regularized PriorBand for Efficient Joint HPO and NAS}
\label{sec:approach:hpo_nas}

Building upon the flexible \autonnunet{} framework, we further enhance the optimization process using \textbf{Regularized PriorBand}.
In this section, we describe how we extend PriorBand~\citep{mallik-neurips23a} from HPO to JAHS.
Given that \nnunet{} requires considerable training cost and provides a strong prior configuration, we aim to incorporate this knowledge into the optimization process to improve its efficiency.
To achieve this, we leverage PriorBand~\citep{mallik-neurips23a}, a multi-fidelity HPO method specifically designed to integrate prior knowledge into the optimization of computationally expensive deep learning models.
It enhances exploration by combining random, prior-based, and incumbent-based sampling strategies to dynamically adjust as the optimization progresses.
Random sampling explores the search space, prior-based sampling leverages expert knowledge, and incumbent-based sampling refines the current best-performing configuration.

We extend the HPO search space of PriorBand by encoding architectures within a unified configuration space~\citep{zela-icml18a}.
However, exploring larger models introduces computational challenges.
While increased model size can enhance accuracy, it also raises optimization costs and prolongs training.
We consider training runtime as an optimization objective to better reflect the practical constraints of medical environments, where computational resources are often limited and large-scale training may be infeasible~\citep{rayed-infmed24}.
Dataset heterogeneity — due to technical factors (e.g., scanners, protocols) and anatomical variability (e.g., organ shape, number of structures) — often necessitates repeated fine-tuning or model adaptation.
In such continual learning scenarios, where retraining is recurring and costly~\citep{isensee-nature19, wagner-corr-2024}, efficient training is essential.
Privacy constraints often prevent centralized access to patient data, requiring localized or federated retraining when new data becomes available, which further emphasizes the importance of minimizing training costs~\citep{wagner-corr-2024}.

The central idea of Regularized PriorBand is that larger models should only be considered if they contribute to accuracy improvement.
Ultimately, the goal remains to optimize for accuracy, ensuring that the best-performing configurations are not discarded in favor of resource-constrained choices. 
An overview of Regularized PriorBand is provided in Algorithm~\ref{alg:approach:reg_priorband} in Appendix~\ref{sec:appendix:approach}.

\paragraph{Selection Strategy in Successive Halving}
A key adaptation in Regularized PriorBand involves modifying the configuration selection strategy in the Successive Halving (SH) subroutine~\citep{jamieson-aistats16a}.
In the standard SH approach, configurations for the next higher budget are selected based on their cost, with the configurations exhibiting the lowest cost being prioritized for evaluation.
However, when optimizing for both accuracy and training runtime, we must consider a vector of costs rather than a single scalar to account for both objectives.

To integrate both accuracy and runtime, we modify the selection process by employing non-dominated sorting and crowding distance sorting as proposed by \citet{deb-ieee02} and similar to recent work~\citep{izquierdo-icml21a,schmucker-arxiv21,awad-arxiv23}.
After evaluating configurations at the current budget, we apply non-dominated sorting to group configurations into fronts. 
To favor diverse solutions, we rank the configurations by their crowding distance within each front.
From these sorted fronts, we select the top $k$ configurations for evaluation at the next higher budget, beginning with the first front and continuing until $k$ configurations are chosen.
If two configurations have equal crowding distances, the selection prioritizes accuracy.
This guarantees that the configuration with the highest accuracy is always promoted.

\paragraph{Incumbent Selection}
In Regularized PriorBand, the final incumbent configuration is selected based on accuracy, without considering runtime.
However, to enable incumbent-based sampling throughout the optimization, we incorporate both accuracy and runtime.
The selection is limited to configurations on the approximated Pareto front, ensuring a balance between the two objectives.
To choose the incumbent for the local search, we compute the area spanned by the normalized objective costs and select the configuration that maximizes this area, facilitating the exploration of trade-offs between accuracy and runtime.

\section{Experimental Setup}
\label{sec:experiments}

Based on our \autonnunet{} framework, we conduct the most comprehensive study on AutoML for MIS to date, being equivalent to approximately $60\text{ k}$ GPU hours and $10\;964 \text{ kg CO}_2$ equivalents (see Appendix~\ref{sec:appendix:hardware} for more details).
Instructions to reproduce all experiments, results, and visualizations can be found in our GitHub repository at \url{https://github.com/automl/AutoNNUnet}.
All experiments are performed using 5-fold cross-validation.
See Appendix~\ref{sec:appendix:experiments:pipeline} for more details.

\subsection{Datasets}
\label{sec:experiments:datasets}

To ensure a comprehensive evaluation of our methods, we use the Medical Segmentation Decathlon (MSD)~\citep{simpson-arxiv19,antonelli-nature22}, a benchmark of ten MIS datasets designed to capture diversity across clinical tasks, imaging modalities, and data characteristics (see Appendix~\ref{sec:appendix:datasets}).
The MSD uses the Dice Similarity Coefficient~(DSC)~\citep{dice_ecology45}, an effective metric for evaluating MIS methods~\citep{zijdenbos_ieee94}.
The DSC measures the overlap between ground truth (X) and prediction (Y) as
$ \text{DSC}(X,Y) = \frac{2|X\cap Y|}{|X| + |Y|}$
ranging from $0$ (no overlap) to $1$ (perfect overlap).

\subsection{Baselines}
\label{sec:experiments:baselines}
In our experiments, we aim to investigate how AutoML methods can improve the segmentation performance of current MIS methods.
Our first baseline is the 3D U-Net of the \nnunet{} framework, particularly its default configurations 
\begin{enumerate*}[label=(\roman*)]
    \item \textit{Conv},
    \item \textit{ResM}, and
    \item \textit{ResL}
\end{enumerate*}.
Additionally, we evaluate \medsamtwo{}~\citep{ma-nature24}, a foundation model-based approach for MIS.
Unlike \nnunet{}, \medsamtwo{} leverages large-scale pre-training and serves as a state-of-the-art competitor to our approach.
We leverage the pipeline proposed by the authors to finetune MedSAM2 on each individual MSD dataset for $100$ epochs, which is roughly equivalent to the training runtime of the most expensive \nnunet{} configuration on \Dds{01}, the dataset with the highest training runtime.

\subsection{Evaluation of \autonnunet{}}
\label{sec:experiments:optimization}

This section outlines the experimental setup for evaluating our \autonnunet{} approach.
\autonnunet{} uses Regularized PriorBand to incorporate the \textit{Training Runtime} objective alongside \textit{\textit{1~-~DSC}} into the optimization to prefer more efficient models at equal performance.
For the PriorBand optimizer, we rely on the setup proposed by \citet{mallik-neurips23a} (see Appendix~\ref{sec:appendix:experiments:automl_methods:priorband}), on one random seed due to the extensive computational resources required otherwise.

The \autonnunet{} search space includes both regular hyperparameters, which define training and configuration settings (e.g., learning rate and data augmentation), and architectural hyperparameters, which govern the network structure (e.g., encoder type and dropout rate). This JAHS-search-space formulation~\citep{bansal-neurips22a} enables simultaneous tuning of training dynamics and model capacity. The full \autonnunet{} search space is given in Table~\ref{tab:experiments:setup:jahs_search_space} and as a combination of the HPO and NAS spaces. Details on the hyperparameters are stated in Appendix~\ref{sec:appendix:experiments:automl_methods:search_spaces}.

\subsection{Ablation Variants} \label{sec:ablation}

To assess the contribution of different components within the \autonnunet{} framework, we define two ablation variants that isolate or modify parts of the JAHS search space:\\
\textbf{HPO using PriorBand.}
In this variant, we disable the architectural search of \autonnunet{} and optimize only the regular hyperparameters, using PriorBand~\citep{mallik-neurips23a}) without the added regularization from \autonnunet{}, to minimize \textit{1 - DSC}.
By excluding architectural hyperparameters, this ablation isolates the effect of tuning configuration choices and helps to quantify the performance gains attributable solely to hyperparameter optimization when the network architecture is fixed to the \nnunet{} default.
In Table~\ref{tab:experiments:setup:jahs_search_space} (top), the ranges and sets of possible values for each hyperparameter are defined.

\textbf{Hierarchical NAS for U-Nets.}
\label{sec:approach:hpo_hnas}
While Regularized PriorBand enables JAHS as in \autonnunet{}, its search space is limited to predefined modifications.
To explore a broader range of U-Net architectures while maintaining efficiency and feasibility, we introduce a hierarchical NAS~(HNAS) search space, leveraging context-free grammars~(CFG)~\citep{schrodi-neurips23a} to systematically refine and optimize U-Net structures (see Appendix~\ref{sec:appendix:approach:hpo_hnas:search_space}).
Our approach preserves \nnunet{}’s default configurations while introducing flexible topological and cell-level design choices.
We apply prior-based sampling with Regularized PriorBand, modeling CFG production rules as categorical and integer hyperparameters to integrate smoothly with existing \nnunet{} components (see Appendix~\ref{sec:appendix:approach:hpo_hnas:prior_based_sampling}).

\section{Results}
\label{sec:results}

In this section, we present empirical results demonstrating the effectiveness of \autonnunet{} across the MSD datasets. 
We evaluate segmentation accuracy, efficiency, and configuration transferability, comparing against baselines and ablations.
To gain a deeper understanding of the underlying optimization behavior, we additionally analyze the importance of individual hyperparameters.

\subsection{\autonnunet{} Results}
\label{sec:results:automl_methods}

\begin{figure}[tb]
    \centering
    \includegraphics[alt={Cost-over-time plots showing trade-offs between segmentation error (1 - DSC [\%], y-axis) and wallclock training time (hours, x-axis) across ten Medical Segmentation Decathlon datasets (D01–D10). Each subplot corresponds to one dataset and compares different methods: nnU-Net (Conv, ResM, ResL), Auto-nnU-Net, HPO, and HPO+HNAS. Lines represent cost over time, and stars indicate final best-performing configurations. Auto-nnU-Net and its variants often achieve lower error compared to nnU-Net baselines.}, width=\linewidth]{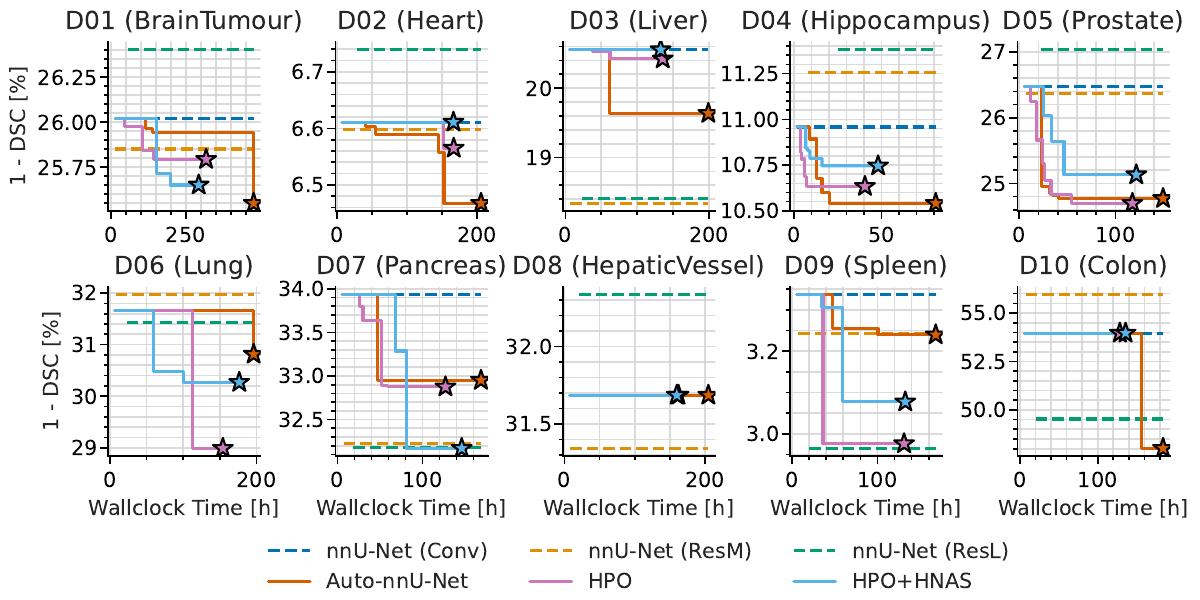}
    \caption{Incumbent performance of \nnunet{}, \autonnunet{}, and \autonnunet{} ablations over time. Detailed results for each dataset are stated in Appendix~\ref{sec:appendix:results}. Final validation DSCs are stated in Table~\ref{tab:results:dsc_overview} in Appendix~\ref{sec:appendix:results}. We exclude \medsamtwo{} as it fails to achieve the performance of \nnunet{} on nine out of ten datasets (see Table~\ref{tab:results:dsc_overview}).}
    \label{fig:results:automl_methods:overview:performance_over_time}
\end{figure}

First, we discuss the optimization progress of \autonnunet{} across MSD datasets.
Figure~\ref{fig:results:automl_methods:overview:performance_over_time} shows the \autonnunet{}s incumbent \odsc{} over time compared to the default \nnunet{} baselines.
Except for \Dds{08}, where the DSC matches, \autonnunet{} outperforms \nnunet{}s convolutional default. 
Notably, for \Dds{04}, \autonnunet{} identifies the incumbent configuration faster than training \resl{}, highlighting its efficiency over computationally expensive models.

The final validation results, including \medsamtwo{}, are stated in Table~\ref{tab:results:dsc_overview} in Appendix~\ref{sec:appendix:results}.
We exclude \medsamtwo{} from the cost-over-time comparison as it underperforms \nnunet{} on nine out of ten datasets, only outperforming it on \Dds{10}.
We hypothesize that \medsamtwo{}s performance on \Dds{10} is due to its requirement of a bounding box prompt from the ground truth mask, which helps with detecting small target regions.
While this approach benefits detection tasks, it requires additional annotation, whereas our method works without such supervision.

\begin{table}[tb]
    \centering
    \begin{tabular}{l|lll|l|ll}
        \toprule
         & \multicolumn{3}{c|}{\textbf{\nnunet{}}} & \makecell[c]{\textbf{Auto-}\\\textbf{nnU-Net}} & \multicolumn{2}{c}{Ablation} \\
        \cmidrule(r@{2pt}){2-4}                        
        \cmidrule(l@{2pt}){6-7}
         & Conv & ResM & ResL & & \makecell{HPO} & \makecell{HPO+HNAS} \\
        \midrule
        D01 & $61.34 \pm 24.3$ & $61.21 \pm 24.0$ & $61.12 \pm 24.4$ & $\mathbf{61.58 \pm 24.3}$ & $61.06 \pm 24.2$ & $57.31 \pm 25.3$ \\
        D02 & $93.33 \pm 1.5$ & $93.36 \pm 1.5$ & $93.03 \pm 1.8$ & $\mathbf{93.46 \pm 1.4}$ & $93.28 \pm 1.5$ & $93.35 \pm 1.5$ \\
        D03 & $85.36 \pm 13.0$ & $86.33 \pm 11.6$ & $\mathbf{86.66 \pm 11.4}$ & $85.91 \pm 12.0$ & $85.36 \pm 13.1$ & $85.68 \pm 12.4$ \\
        D04 & $89.43 \pm 3.8$ & $89.22 \pm 3.8$ & $89.10 \pm 3.7$ & $\mathbf{89.75 \pm 4.0}$ & $88.34 \pm 3.9$ & $88.99 \pm 3.8$ \\
        D05 & $80.91 \pm 7.0$ & $79.98 \pm 7.0$ & $80.65 \pm 6.5$ & $\mathbf{82.29 \pm 5.9}$ & $81.95 \pm 6.4$ & $77.96 \pm 9.3$ \\
        D06 & $67.14 \pm 30.6$ & $62.44 \pm 34.6$ & $\mathbf{70.52 \pm 26.2}$ & $68.78 \pm 26.9$ & $69.70 \pm 26.7$ & $69.83 \pm 21.9$ \\
        D07 & $64.70 \pm 20.5$ & $66.45 \pm 21.0$ & $\mathbf{66.68 \pm 21.3}$ & $65.23 \pm 20.8$ & $66.38 \pm 19.7$ & $65.63 \pm 19.7$ \\
        D08 & $68.37 \pm 19.0$ & $\mathbf{68.48 \pm 18.8}$ & $68.35 \pm 19.2$ & $68.23 \pm 19.2$ & $68.12 \pm 19.3$ & $68.43 \pm 19.1$ \\
        D09 & $97.23 \pm 1.2$ & $97.11 \pm 1.3$ & $94.93 \pm 10.7$ & $97.11 \pm 1.4$ & $\mathbf{97.34 \pm 1.0}$ & $96.62 \pm 1.3$ \\
        D10 & $52.96 \pm 35.4$ & $48.26 \pm 38.2$ & $50.36 \pm 36.2$ & $\mathbf{58.05 \pm 32.0}$ & $47.20 \pm 37.8$ & $50.88 \pm 36.7$ \\
        \midrule
        \textbf{Mean} & $76.08 \pm 15.6$ & $75.28 \pm 16.2$ & $76.14 \pm 16.1$ & $\mathbf{77.04 \pm 14.8}$ & $75.87 \pm 15.4$ & $75.47 \pm 15.1$ \\
        \bottomrule
    \end{tabular}
    \caption{Mean $\pm$ standard deviation of the DSC [\%] for the MSD test set obtained through the official submission platform for all datasets (rows) and approaches (columns). Metrics are computed over all test set instance DCSs per dataset. The best-performing method per dataset is highlighted in \textbf{bold}. Notably, as \medsamtwo{} requires access to the ground truth segmentations to generate prompts, the model cannot be evaluated on unlabeled data.}
    \label{tab:results:dsc_test}
\end{table}

To assess performance on unseen data, we evaluate the MSD test set results.
Table~\ref{tab:results:dsc_test} presents the final test set DSC [\%] for all approaches, excluding \medsamtwo{}, which requires ground truth segmentations that are unavailable for the MSD.
Consistent with the validation results, \autonnunet{} achieves the highest average DSC (77.04\%).
Our method surpasses all \nnunet{} baselines and demonstrates strong generalization, ranking best on five out of ten datasets.

\begin{figure}[htb]
    \centering
    \includegraphics[alt={Qualitative segmentation results for case BRATS_012 in dataset D01 (BrainTumour), showing the best-case performance. Each column corresponds to a method: nnU-Net (Conv), nnU-Net (ResM), nnU-Net (ResL), MedSAM2, Auto-nnU-Net, HPO, and HPO+HNAS, with their respective DSC scores displayed. Rows show axial, sagittal, and coronal slices of the input image, ground truth, and predicted segmentations. Segmentations are overlaid with color-coded tumor classes: yellow for edema, green for non-enhancing tumor, and blue for enhancing tumor. Auto-nnU-Net shows the most accurate and detailed segmentation compared to other methods.}, width=\linewidth]{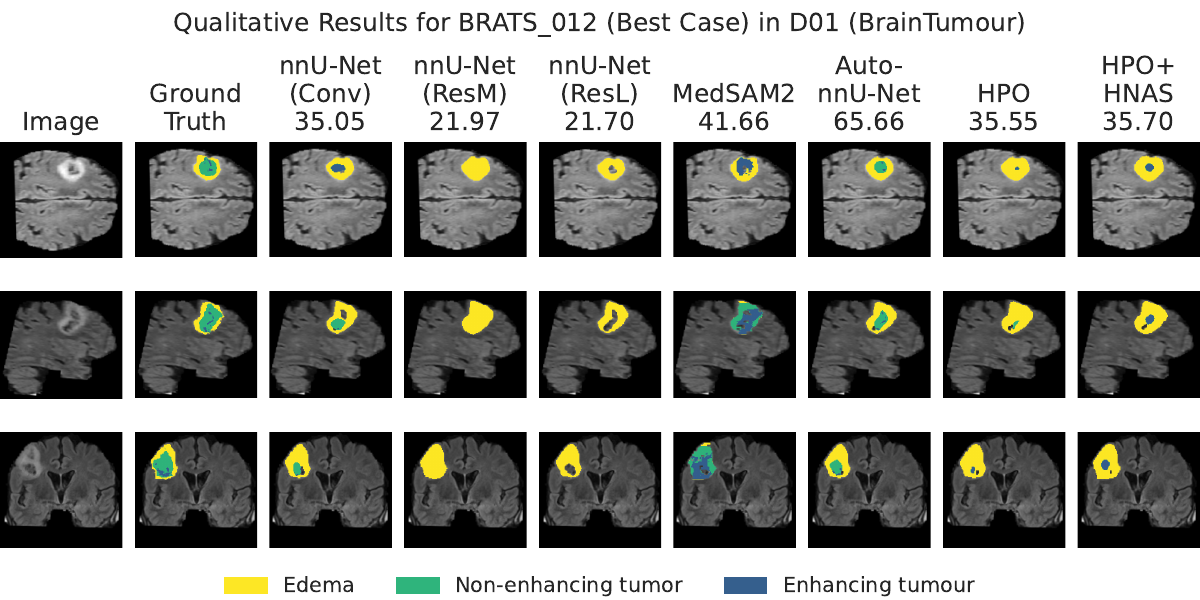}
    \caption{Qualitative segmentation results for \Dds{01}. Columns show the input image, ground truth mask, and method predictions, with colors denoting foreground classes. Numbers below method names indicate DSC scores [\%] for this example. Each row shows a slice of the 3D volume along one axis. As the 4D volume is an mp-MRI scan, the first parameter setting is used to extract a 3D volume. Additional results are in Appendix~\ref{sec:appendix:results}.}
    \label{fig:results:automl_methods:overview:qualitative_d01}
\end{figure}

Figure~\ref{fig:results:automl_methods:overview:qualitative_d01} shows qualitative results for the best validation case in \Dds{01}.
All methods correctly segment the \textit{Edema} class but struggle with the other foreground classes.
Notably, \medsamtwo{} fails to segment the Enhancing tumor voxels within the \textit{Non-enhancing tumor} region and over-segments the \textit{Non-enhancing tumor} class.
In contrast, \autonnunet{} captures fewer voxels of both the \textit{Non-enhancing tumor} and \textit{Enhancing tumor} classes.
These results highlight that \medsamtwo{} over-segments the target regions, while other methods under-segment them.

\begin{figure}[htb]
    \centering
    \includegraphics[alt={Pareto front plots for D03 (Liver) and D04 (Hippocampus) datasets showing the trade-off between segmentation error (1 - DSC [\%], x-axis) and training runtime in hours (y-axis, log scale). Each point represents a method: nnU-Net (Conv, ResM, ResL), MedSAM2, Auto-nnU-Net, HPO, and HPO+HNAS. Lines indicate Pareto-optimal configurations. In both plots, Auto-nnU-Net and especially HPO+HNAS achieve better performance-runtime trade-offs compared to other baselines, with notably lower runtimes at competitive or improved accuracy.}, width=\textwidth]{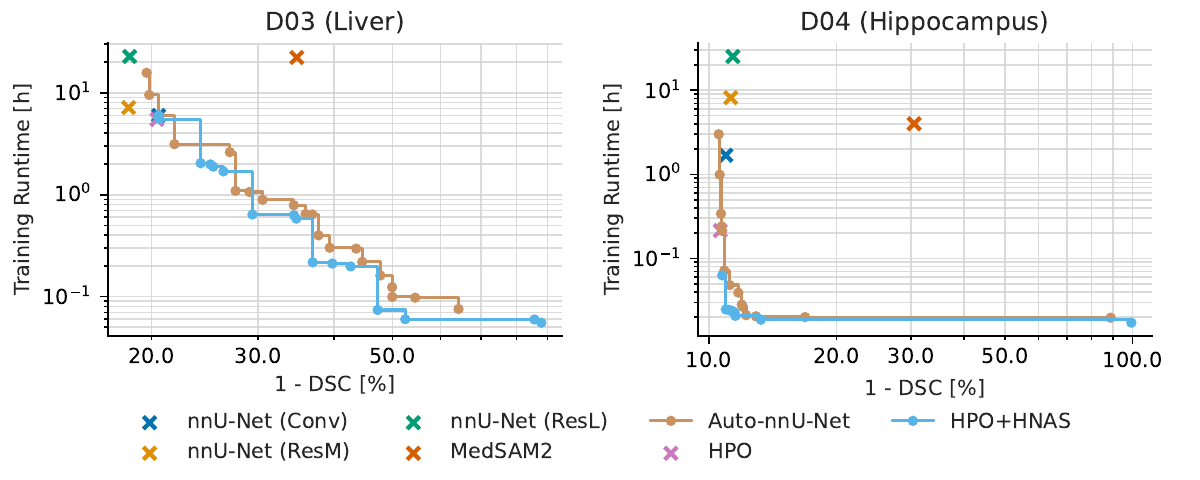}
    \caption{Pareto fronts of \autonnunet{} and \hpohnas{} for \Dds{03} \textbf{(left)} and \Dds{04} \textbf{(right)} compared to the baselines and \hpo{} results. Additional results are stated in Appendix~\ref{sec:appendix:results}.}
    \label{fig:results:pareto_fronts:two_datasets}
\end{figure}

Regularized PriorBand inherently balances accuracy and training runtime during optimization (Section~\ref{sec:approach:hpo_nas}).
Figure~\ref{fig:results:pareto_fronts:two_datasets} compares the Pareto fronts of \autonnunet{}, its ablations (\hpo{}, \hpohnas{}), and baselines (\nnunet{}, \medsamtwo{}) on \Dds{03} and \Dds{04}, illustrating objective trade-offs.
On \Dds{03}, \autonnunet{} and \hpohnas{} reveal clear accuracy-runtime trade-offs, while \resm{} achieves high DSC with low runtime.
On \Dds{04}, \autonnunet{} and \hpohnas{} outperform all \nnunet{} variants in accuracy and significantly reduce training time — \hpohnas{} cuts runtime by a factor of 26.
\medsamtwo{} underperforms on both datasets.
These results demonstrate \autonnunet{}’s ability to jointly optimize accuracy and efficiency.

\paragraph{Ablation Results}
We discuss the \hpo{} and \hpohnas{} ablations of \autonnunet{} in Figure~\ref{fig:results:automl_methods:overview:performance_over_time} and Table~\ref{tab:results:dsc_test}.
\hpo{} outperforms \conv{} on all datasets except \Dds{08} and \Dds{10}, while \autonnunet{} shows similar improvements, excelling on all but \Dds{08}.
\hpohnas{} surpasses \conv{} on six datasets and generally optimizes more efficiently than both \autonnunet{} and \hpo{} for some datasets, like \Dds{01}.
However, compared to \autonnunet{}, both \hpo{} and \hpohnas{} exhibit lower DSCs on the test set, suggesting reduced robustness to unseen data.
For \Dds{01}, despite similar validation DSCs, \hpohnas{} underperforms relative to \autonnunet{}, indicating greater sensitivity to unseen data and effectiveness of encoding neural architectures as hyperparameters.

\subsection{Analysis of Hyperparameter Importance in \autonnunet{}}
\label{sec:results:automl_analysis}

To assess the impact of individual hyperparameters on accuracy, we use functional ANOVA~(fANOVA)~\citep{hutter-icml14a} to estimate their global importance across the configuration space by decomposing performance variance into contributions from each hyperparameter and their interactions.
Figure~\ref{fig:results:automl_analysis:hpo:global_hpis} shows their importance across datasets.
Key hyperparameters such as \textit{Foreground Oversampling}, \textit{Initial Learning Rate}, and \textit{Momentum (SGD)} are consistently influential, though importance varies notably between datasets.
This highlights the value of AutoML over fixed settings, as used in the original \nnunet{}.
In contrast, hyperparameters like \textit{Encoder Type} and \textit{Normalization} show low importance and may not require further optimization.

\subsection{Transferring Incumbent Configurations across Datasets}
\label{sec:results:transferring_configs}

A key question in evaluating dataset influence on AutoML is whether an optimized configuration for one dataset generalizes to others.
We analyze the transferability of \autonnunet{}s incumbents, excluding \Dds{08} where it does not outperform the default, resulting in a $9 \times 10$ matrix (Figure~\ref{fig:appendix:results:automl_results:cross_eval_matrix}).
In half of the datasets, the tailored incumbent does not achieve the highest DSC, with the \Dds{02} configuration showing the largest gain (+2.78\%) on \Dds{05}, while the \Dds{03} incumbent achieves the highest DSC on four datasets.
However, its lower performance on \Dds{05} prevents it from surpassing the \nnunet{} default on average.
These results suggest that configurations can transfer across datasets — e.g., \Dds{03} performs best on \Dds{03}, \Dds{08}–\Dds{10} — but others, like \Dds{04} and \Dds{05}, generalize poorly, particularly to \Dds{06}–\Dds{10}.
This highlights the potential of meta-learned HPO for improved transferability in MIS~\citep{feurer-aaai15a,wistuba-dsaa15a,schilling-pkdd16a}.

\section{Conclusion and Future Work}
\label{sec:discussion}

In this work, we proposed \autonnunet{}, an automated framework for medical image segmentation that combines \nnunet{} with structured HPO and NAS.
By integrating Regularized PriorBand, we jointly optimize segmentation performance and training runtime, addressing practical constraints in medical settings.
Our comprehensive evaluation on all ten Medical Segmentation Decathlon datasets demonstrates that \autonnunet{} consistently outperforms or matches strong baselines while maintaining practical resource requirements.
We further analyzed the contributions of HPO and NAS through ablation studies, examined the transferability of optimized configurations across datasets, and assessed hyperparameter importance. These insights contribute to a deeper understanding of the design and optimization of our segmentation approach in diverse clinical settings.
Overall, \autonnunet{} provides a flexible and resource-aware foundation for automated medical image segmentation, enabling robust model design under real-world constraints.

\paragraph{Limitations}
This study, including results for both \nnunet{} and \autonnunet{}, is based on the 3D U-Net architecture without post-processing or ensembling, which may not fully reflect the original \nnunet{}’s performance~\citep{isensee-nature19}.
However, incorporating ensembling — common in AutoML~\citep{erickson-arxiv20a} — would likely enhance \autonnunet{}’s results.
Regarding our Pareto analysis, lower-budget configurations approximate full-budget performance and reveal runtime-accuracy trade-offs.
Lastly, while surrogate models in DeepCAVE may introduce slight approximation errors, the findings provide a strong foundation for advancing AutoML in MIS.

\paragraph{Future research}
Future work could extend evaluations to the full \nnunet{} pipeline and further investigate how dataset properties affect AutoML outcomes.
Warm-starting AutoML with multiple default configurations~\citep{pfisterer-arxiv18a} and meta-learning~\citep{feurer-aaai15a,vanschoren-automlbook19a,aguiar-prl19} could improve the efficiency of AutoML for MIS.
Finally, zero-shot AutoML with pre-trained models~\citep{ferreira-icml22a} could enhance adaptability while reducing costs.

\section{Broader Impact Statement}
\label{sec:results:broader_impact_statement}
AutoML for MIS can improve diagnostic accuracy and efficiency by reducing manual tuning and supporting advanced model development in collaboration with medical professionals — making it more accessible to institutions with limited ML expertise.
Accurate segmentation aids early diagnosis and treatment planning, while efficient optimization is crucial in resource-constrained settings.
Challenges remain, including performance dependence on training data and potential bias, which can hinder generalization. AutoML can help mitigate this by reducing expert dependence and enabling optimization across diverse datasets, promoting fairness.
Future work should focus on fairness-aware methods and more efficient AutoML strategies to support ethical, sustainable deployment.
If these challenges are addressed, AutoML could become a powerful tool for MIS, improving diagnostic robustness and precision while ensuring ethical and responsible deployment.

\begin{acknowledgements}
  This work was supported by the Federal Ministry of Education and Research (BMBF), Germany, under the AI service center KISSKI (grant no. 01IS22093C).
  Leona Hennig and Marius Lindauer acknowledge funding by the German Federal Ministry of the Environment, Nature Conservation, Nuclear Safety and Consumer Protection (GreenAutoML4FAS project no. 67KI32007A).
  Steffen Oeltze-Jafra and Marius Lindauer were supported by the Lower Saxony Ministry of Science and Culture (MWK) with funds from the Volkswagen Foundation's zukunft.niedersachsen program [project name: CAIMed - Lower Saxony Center for Artificial Intelligence and Causal Methods in Medicine; grant number: ZN4257].
\end{acknowledgements}

\printbibliography[nottype=online]

\appendix

\newpage
\section{Hardware and Resource Consumption}
\label{sec:appendix:hardware}

Compute nodes were equipped with the following software and hardware:
\begin{itemize}
    \item \textbf{OS}: Rocky Linux 9.5
    \item \textbf{CPU}: 48xAMD EPYC 9354 32-Core Processor
    \item \textbf{RAM}:120 GB
    \item \textbf{GPU}: 1xNVIDIA H100 PCIe, 80GB VRAM, CUDA 12.4
\end{itemize}

All experiments took a total of $59\;945$ GPU hours, with an estimated power consumption of $0.5 \text{ kWh}$ per GPU hour.
This results in a total power consumption $29\;972.5 \text{ kWh}$ and $10\;964 \text{ kg CO}_2$ equivalents based on the average energy mix of Germany in 2023\footnote{\url{https://tco2e.net/kwh/country/germany/}}.

\section{Related Work}
\label{sec:appendix:related}

\subsection{PriorBand}
\label{sec:appendix:related:priorband}

In MIS, architectures and hyperparameter settings designed by experts can serve as good starting points for HPO, and leveraging this knowledge may accelerate the optimization, particularly given the high computational demands of training deep learning models.
However, although BO and BOHB improve the performance of HPO, they do not explicitly incorporate this knowledge.

PriorBand~\citep{mallik-neurips23a} addresses this limitation by integrating prior configurations, e.g., expert beliefs, into the optimization process and improves the anytime performance of existing methods such as $\pi$BO~\citep{hvarfner-iclr22a}.
By incorporating prior knowledge about well-performing regions in the search space, PriorBand aims to enhance the efficiency of HPO in computationally expensive scenarios with a strong prior configuration.
As the work presented in this thesis largely relies on PriorBand, we provide a more detailed discussion of this approach.

\begin{figure}[tb]
\begin{algorithm}[H]
\caption{PriorBand HPO algorithm \citep{mallik-neurips23a}}
\label{alg:related:priorband}
\KwIn{Budgets $b_\text{min}$ and $b_\text{max}$, reduction factor $\eta$ (default $\eta = 3$), prior $\pi$}
\KwOut{Incumbent configuration $\lambda^*$}
\SetKwFunction{Update}{Update}
\SetKwProg{Fn}{Function}{:}{}
$\mathcal{H} \gets \emptyset$ \tcp*{all observations}
$s_\text{max} = \lfloor \log_\eta \frac{b_\text{max}}{b_\text{min}} \rfloor$\;
\For{$s \in \{s_\text{max}, s_\text{max} - 1, \dots, 0\}$}{
    $r \gets s_\text{max} - s$\;
    $p_\mathcal{U} \gets 1 / (1 + \eta^{r})$\;
    $p_\pi \gets  1 - p_\mathcal{U}$\;
    $p_{\hat \lambda} \gets 0$\;
    \If{evaluated at least one config at $b_\text{max}$}{
        $p_\pi, p_{\hat \lambda} \gets$ \textsc{DynamicWeighting}($\mathcal{H}, r, p_\pi$)\;
    }
    $n \gets \lceil \frac{s_\text{max}}{s + 1} \rceil$ configurations\;
    \For{$i \in \{1, \dots, n\}$}{
         $d(\cdot) \gets$ sample strategy by $\{p_\mathcal{U}, p_\pi, p_{\hat \lambda}\}$\;
         $\lambda_i \gets$ sample from $d(\cdot)$\;
    }
    Run successive halving on the configurations $\lambda_i$ with initial budget $\eta^{-s} \cdot b_\text{max}$\;
    Add observations to $\mathcal{H}$\;
    $\lambda^* \gets \text{arg}\min_{(\lambda, c) \in \mathcal{H}}{} \text{ } c$\;
}
\Return $\lambda^*$\;
\end{algorithm}
\end{figure}

An outline of PriorBand is shown in Algorithm~\ref{alg:related:priorband}.
PriorBand replaces the random sampling in HB with an ensemble sampling strategy $\mathcal{E}_\pi$ (see lines 12-13), containing the following three components:
\begin{enumerate}[label=(\roman*)]
    \item \textbf{Random Sampling from $\mathcal{U}(\cdot)$}. This strategy samples hyperparameter configurations uniformly from the search space. It enables exploration of the configuration space to find promising regions without relying on the prior distribution.
    \item \textbf{Prior-based Sampling from $\mathcal{\pi}(\cdot)$}. This sampling strategy leverages expert knowledge about well-performing configurations.
    It facilitates a local search near the prior configuration using perturbation. If the preceding configuration is accurate, it accelerates the optimization process.
    \item \textbf{Incumbent-based Sampling from $\hat \lambda(\cdot)$}. This strategy samples configurations around the current best-performing configuration. By exploring the configuration space locally around the incumbent configuration, it aims to refine and improve upon it.
    This strategy is beneficial if the prior is not accurate or useful.
\end{enumerate}

Each of the three sampling components is assigned a weight that determines the probability of the respective strategy being used when sampling from $\mathcal{E}_\pi$.
The weights are denoted as $p_\mathcal{U}$, $p_\pi$, and $p_{\hat \lambda}$.
Initially, PriorBand assigns equal weights to random and prior-based sampling to ensure a balance between exploration and leveraging expert knowledge.
As the optimization progresses, the probability of random sampling decreases geometrically, increasing the proportion of the other two strategies (see line 5 of Algorithm~\ref{alg:related:priorband}).

Once the first configuration is evaluated on the maximum budget, prior-based and incumbent-based sampling are weighted dynamically (see lines 8-9 of Algorithm~\ref{alg:related:priorband}).  
In the \textsc{DynamicWeighting} subroutine, configurations are ranked by their performance, and the likelihood of the top configurations under the prior and incumbent distributions is computed.
Based on these likelihoods, PriorBand dynamically adjusts the sampling weights, assigning a higher weight to the distribution that is more likely to produce well-performing configurations.
The weighting ensures efficiency in the case of well-performing and robustness against bad prior configurations.
Using the weights $p_\mathcal{U}$, $p_\pi$, and $p_{\hat \lambda}$, a sampling strategy is selected for each configuration individually.
Based on the chosen strategy, a hyperparameter configuration is sampled.
Subsequently, similar to HB, SH is called as a subroutine to efficiently allocate resources to the most promising configurations.

\paragraph{Prior-based Sampling}
In PriorBand, prior-based sampling of hyperparameters is accomplished by drawing samples from a prior distribution.
The type and shape of the distribution are determined by
\begin{enumerate*}[label=(\roman*)]
    \item the type of hyperparameter and
    \item the prior confidence provided by the user.
\end{enumerate*}
For numerical hyperparameters, a truncated normal distribution is defined over the range of possible values, where the mean is set to the default value.
The prior confidence adjusts the standard deviation, with higher confidence resulting in a lower standard deviation.
Integer hyperparameters are sampled by rounding the values to the nearest integer value.
For categorical hyperparameters, in contrast, the probabilities are uniformly distributed across all values except for the default, whose probability is increased in proportion to the prior confidence.

\newpage
\section{Approach}
\label{sec:appendix:approach}

\subsection{\autonnunet{}}
\label{sec:appendix:approach:autonnunet}

Our framework extends \nnunet{} by
\begin{enumerate*}[label=(\roman*)]
    \item the \texttt{AutoExperimentPlanner},
    \item the \texttt{CFGUNet}, and
    \item the \texttt{AutoNNUNetTrainer}.
\end{enumerate*}
Figure~\ref{fig:related:autonnunet_overview} shows an overview of the \autonnunet{} framework.
Unlike \nnunet{}, which relies on fixed and rule-based configurations, \autonnunet{} introduces a hyperparameter configuration $\lambda$ and architecture $A$ as inputs alongside the dataset fingerprint.
This enables a more flexible and automated experiment planning and training process, allowing models to be systematically optimized based on different hyperparameter and architecture choices.
Then, \autonnunet{} returns the validation accuracy and training runtime as objectives.
Our interface enables powerful and flexible search strategies, including multi-objective optimization, where trade-offs between performance and efficiency can be explicitly modeled.

\autonnunet{} extends \nnunet{} with three key components that enable flexible integration of AutoML methods.
The \texttt{AutoExperimentPlanner} incorporates hyperparameter configurations into the planning process to enable the optimization of architectural properties such as the number of features, normalization, activation, and dropout.
To support more expressive architectural definitions through hierarchical NAS search spaces (see Section~\ref{sec:approach:hpo_hnas}), the \texttt{CFGUNet} translates function composition representations into neural network models.
Finally, the \texttt{AutoNNUNetTrainer} extends the training pipeline with dynamic hyperparameter configurations, including optimizer settings, learning rate schedules, and augmentation strategies.
Together, these components provide a unified framework for optimizing both hyperparameters and architectures in MIS.

\subsection{Regularized PriorBand}
\label{subsec:appendix:regpriorband}

An overview of the Regularized PriorBand algorithm is shown in Algoritm~\ref{alg:approach:reg_priorband}, where changes compared to the original PriorBand algorithm (see Algorithm~\ref{alg:related:priorband}) are highlighted.
For a detailed outline of PriorBand, we refer to Appendix~\ref{sec:appendix:related:priorband}.
In Line 21, we apply non-dominated sorting on the set of observations, i.e., candidate configurations $P$ in the current stage of SH.
The subroutine returns a list of fronts, where the first front is the actual Pareto front of $P$ and each subsequent the updated Pareto front after removing the previous front.
In Lines 23-24, we iterate over all fronts and sort the configurations within the front based on their 
\begin{enumerate*}[label=(\roman*)]
    \item crowding distance and
    \item objective cost
\end{enumerate*}
by calling the \textsc{crowdingDistanceAndAccuracySorting} subroutine.
It sorts the configurations descendingly based on their crowding distance and, in case of equal crowding distances, the cost of the primary objective, e.g., accuracy.
Regularized PriorBand thereby only considers the primary objective when a Pareto front consists only of two points with equal crowding distance.

\begin{figure}[h]
\begin{algorithm}[H]
\caption{Regularized PriorBand}
\label{alg:approach:reg_priorband}
\KwIn{Budgets $b_\text{min}$ and $b_\text{max}$, reduction factor $\eta$ (default $\eta = 3$), prior configuration $\pi$}
\KwOut{Incumbent configuration $\lambda^*$}
\SetKwFunction{Update}{Update}
\SetKwProg{Fn}{Function}{:}{}

$\mathcal{H} \gets \emptyset$ \tcp*{all observations}
$s_\text{max} = \lfloor \log_\eta \frac{b_\text{max}}{b_\text{min}} \rfloor$\;
\tcp{HyperBand}
\For{$s \in \{s_\text{max}, s_\text{max} - 1, \dots, 0\}$}{
    $r \gets s_\text{max} - s$\;
    $p_\mathcal{U} \gets 1 / (1 + \eta^{r})$\;
    $p_\pi \gets  1 - p_\mathcal{U}$\;
    $p_{\hat \lambda} \gets 0$\;
    \If{evaluated at least one config at $b_\text{max}$}{
        $p_\pi, p_{\hat \lambda} \gets$ \textsc{DynamicWeighting}($\mathcal{H}, r, p_\pi$)\;
    }
    \tcp{Sampling configurations}
    $n \gets \lceil \frac{s_\text{max}}{s + 1} \rceil$ configurations\;
    \For{$i \in \{1, \dots, n\}$}{
         $d(\cdot) \gets$ sample strategy by $\{p_\mathcal{U}, p_\pi, p_{\hat \lambda}\}$\;
         $\lambda_i \gets$ sample from $d(\cdot)$\;
    }
    \begingroup
    \color{changedcolor}
    \tcp{Successive halving~(SH)}
    $\mathcal{C} \gets [\lambda_1, \dots, \lambda_n]$\;
    $k \gets \frac{n}{\eta}$
    \tcp*{Number of configurations for next stage}
    \For{$b \in \{\eta^{-s} \cdot b_\text{max}, \eta^{-(s - 1)} \cdot b_\text{max}, \dots, b_\text{max}\}$}{
        $P \gets \emptyset$
        \tcp*{Candidates for current stage in SH}
        \For{$\lambda \in \mathcal{C}$}{
            $c \gets$ \textsc{Evaluate}($\lambda$, $b$)
            \tcp*{Evaluate and return cost vector}
            $P \gets P~\cup\{(\lambda,c)\}$\;
        }
        
    $F_1, \dots, F_m \gets$ \textsc{NonDominatedSorting}$(P)$\;
    $\mathcal{C} \gets [~]$\;
    \For{$F \in \{F_1, \dots, F_m\}$}{
        \tcp{We (1) sort based on crowding distance descendingly and (2) based on 1 - DSC ascendingly}
        $\mathcal{C} \gets \mathcal{C}~+~$\textsc{CrowdingDistanceAndCostSorting}$(F)$\;
            }
        $\mathcal{C} \gets [\mathcal{C}_1, \dots, \mathcal{C}_k$]
        \tcp*{Take $k$ best candidates}
        $k \gets \frac{k}{\eta}$\;
        $\mathcal{H} \gets \mathcal{H}~\cup P$\;
    }
    $\hat \lambda \gets$ \textsc{GetIncumbent}($\mathcal{H}$)\;
    $\lambda^* \gets \text{arg}\min_{(\lambda, c) \in \mathcal{H}}{} \text{ } c_0$\;
    \endgroup
}
\Return $\lambda^*$\;
\end{algorithm}
\end{figure}

\subsection{Hierarchical NAS for U-Nets}
\label{sec:appendix:approach:hpo_hnas}

\subsubsection{Search Space}
\label{sec:appendix:approach:hpo_hnas:search_space}

In this section, we describe the construction of our hierarchical neural architecture search space using a context-free grammar~(CFG) based on the work of \citet{schrodi-neurips23a}.
Additionally, we extract architecture-level features as numerical and categorical pseudo-hyperparameters, reflecting architectural properties, from the function composition representation.
This facilitates post-hoc analyses, offering insights into how design choices affect segmentation performance.

Since the space of allowed architectures is constrained by the shape of the input images, we dynamically generate the context-free grammar~(CFG) tailored to the dataset at hand.
To determine the maximum number of stages, i.e., the possible number of downsampling operations, we leverage \nnunet{}s experiment planning framework.
It iteratively computes the downsampled image size until the minimum feature map size of $4 \times 4 \times 4$ voxels is reached.
We refer to this number as $n_{\text{stages,max}}$.
Table~\ref{tab:appendix:approach:search_space_sizes} shows the search space sizes for different values of $n_{\text{stages, max}}$.

\begin{table}[t]
    \centering
    \begin{tabular}{rr}
        \toprule
        $n_\text{stages,max}$ & Search Space Size \\
        \midrule
        4 & $502\,400$ \\
        5 & $2\,140\,800$ \\
        6 & $8\,678\,400$ \\
        7 & $34\,892\,800$ \\
        \bottomrule
    \end{tabular}
	\caption{Hierarchical NAS search space sizes based on the maximum number of stages determined by nnU-Net. Sizes are computed following the method proposed by \citet{schrodi-neurips23a}.}
    \label{tab:appendix:approach:search_space_sizes}
\end{table}

We begin with the starting symbol \texttt{S}.
The first production rule specifies the number of stages $n_{\text{stages}}$ in the U-Net, which can take values in the range $\left[ \left\lfloor \frac{n_{\text{stages,max}}}{2} \right\rfloor, n_{\text{stages,max}}\right]$.
For example, if $n_{\text{stages,max}} = 4$, the first production rule is defined as
\begin{align}
    S~::=~&~\texttt{U-Net}(2E,\;2D)~~|~~\texttt{U-Net}(3E,\;3D)~~|~~\texttt{U-Net}(4E,\;4D) \quad ,
    \label{eq:approach:hnas:starting_rule}
\end{align}
where \texttt{U-Net} is a terminal symbol.
The nonterminal symbols $2E, \dots, 4E$ and $2D, \dots, 4D$ represent encoder and decoder modules of two, three, and four stages, respectively.

For the encoder the following production rules determine whether to use a convolutional or residual encoder:
\begin{equation}
    \label{eq:approach:hnas:encoder_rule}
    \begin{aligned}
        2E~::=~&~\texttt{ConvEncoder}\left(E_{\text{Norm}} E_{\text{Nonlin}} E_{\text{Dropout}}, CEB_1, \texttt{down}, CEB_2 \right)~| \\
        &~\texttt{ResEncoder}\left(E_{\text{Norm}} E_{\text{Nonlin}} E_{\text{Dropout}}, REB_1, \texttt{down}, REB_2 \right) \\
        3E~::=~&~\texttt{ConvEncoder}\left(E_{\text{Norm}} E_{\text{Nonlin}} E_{\text{Dropout}}, CEB_1, \texttt{down}, \dots, CEB_3 \right)~| \\
        &~\texttt{ResEncoder}\left(E_{\text{Norm}} E_{\text{Nonlin}} E_{\text{Dropout}}, REB_1, \texttt{down}, \dots, REB_3 \right) \\
        4E~::=~&~\texttt{ConvEncoder}\left(E_{\text{Norm}} E_{\text{Nonlin}} E_{\text{Dropout}}, CEB_1, \texttt{down}, \dots, CEB_4 \right)~| \\
        &~\texttt{ResEncoder}\left(E_{\text{Norm}} E_{\text{Nonlin}} E_{\text{Dropout}}, REB_1, \texttt{down}, \dots, REB_4 \right) \quad .
    \end{aligned}
\end{equation}
The terminal symbols \texttt{ConvEncoder} and \texttt{ResEncoder} correspond to the respective nnU-Net building blocks, while the terminal symbol \texttt{down} represents the downsampling operation.
Depending on the type of encoder, a sequence of convolutional encoder or residual encoder blocks is introduced.
For each stage $i \in [1,n_{\text{stages}}]$, they are denoted by the nonterminals $CEB_i$ and $REB_i$ for a convolutional and residual encoder, respectively.
Additionally, the nonterminal symbols $E_{\text{Norm}}$,  $E_{\text{Nonlin}}$, and $E_{\text{Dropout}}$ are introduced to represent normalization, non-linearity, and dropout components.

Similar to the encoder, the decoder production rules are constructed, but with only one type of decoder:
\begin{equation}
    \label{eq:approach:hnas:decoder_rule}
    \begin{aligned}
        2D~::=~&~\texttt{ConvDecoder}\left(D_{\text{Norm}} D_{\text{Nonlin}} D_{\text{Dropout}}, \texttt{up}, DB_1 \right) \\
        3D~::=~&~\texttt{ConvDecoder}\left(D_{\text{Norm}} D_{\text{Nonlin}} D_{\text{Dropout}}, \texttt{up}, DB_1, \texttt{up}, DB_2 \right) \\
        4D~::=~&~\texttt{ConvDecoder}\left(D_{\text{Norm}} D_{\text{Nonlin}} D_{\text{Dropout}}, \texttt{up}, DB_1, \dots, DB_3 \right) \quad .
    \end{aligned}
\end{equation}
Here, the nonterminals \texttt{ConvDecoder} with its corresponding decoder blocks $DB_i$ for stages $i \in [1,n_{\text{stages}} - 1]$ are introduced.
We note that the last encoder block with index $n_{\text{stages}}$ represents the bottleneck.
Thus, the decoder contains one fewer block than the encoder.

With the production rules introduced so far, we can define both the overall topology and encoder type of the U-Net.
To specify the actual number of blocks per stage, the nonterminals are replaced with terminal symbols representing the block count.
The possible block counts for each stage are derived from \nnunet{}s default configuration.
Depending on the encoder type, each stage has a fixed number of blocks, denoted as $n_{\text{CEB}, i}$ and $n_{\text{REB}, i}$ for the convolutional and residual encoders, respectively.
Similarly, the number of blocks per stage in the decoder is denoted as $n_{\text{DB}, i}$.
To control the overall model size, we introduce a maximum model scale $S_\text{max}$.
This leads to the following production rules:
\begin{equation}
    \label{eq:approach:hnas:block_rules}
    \begin{aligned}
        CEB_i~&~::=~~\texttt{1b}~~|~~\texttt{2b}~~|~~\dots~~\{S_\text{max} \cdot n_{\text{CEB},i}\}\texttt{b} \\
        REB_i~&~::=~~\texttt{1b}~~|~~\texttt{2b}~~|~~\dots~~\{S_\text{max} \cdot n_{\text{REB},i}\}\texttt{b} \\
        DB_i~&~::=~~\texttt{1b}~~|~~\texttt{2b}~~|~~\dots~~\{S_\text{max} \cdot n_{\text{DB},i}\}\texttt{b} \quad .
    \end{aligned}
\end{equation}
The terminal symbols \texttt{1b}, \texttt{2b}, \dots represent the number of blocks in the respective stage, with $\{S_\text{max} \cdot n_{\text{CEB}, i}\}$ acting as a placeholder that is replaced when the CFG is constructed.

To balance search space size and expressiveness, we allow different normalization, non-linearity, and dropout configurations for the encoder and decoder.
These are defined by the following production rules:
\begin{equation}
    \label{eq:approach:hnas:norm_nonlin_dropout_rules}
    \begin{aligned}
          E_{\text{Norm}}, D_{\text{Norm}}~::=~&~\texttt{InstanceNorm}~~|~~\texttt{BatchNorm} \\
        E_{\text{Nonlin}},D_{\text{Nonlin}}~::=~&~\texttt{LeakyReLU}~~|~~\texttt{ReLU}~~|~~\texttt{ELU}~~|~~\texttt{PReLU}~~|~~\texttt{GELU} \\
            E_{\text{Dropout}}, D_{\text{Dropout}}~::=~&~\texttt{Dropout}~~|~~\texttt{NoDropout} \quad .
    \end{aligned}
\end{equation}

Here, we state an examplary search space for $n_{\text{stages,max}} = 4$ and $S_{\text{max}} = 2$:

\begin{equation}
\label{eq:appendix:approach_details:example_search_space}
    \resizebox{0.7\linewidth}{!}{
        $\begin{aligned}
            S~::=~&~\texttt{U-Net}(2E,\;2D)~~|~~\texttt{U-Net}(3E,\;3D)~~|~~\texttt{U-Net}(4E,\;4D) \\
            2E~::=~&~\texttt{ConvEncoder}\left(E_{\text{Norm}} E_{\text{Nonlin}} E_{\text{Dropout}}, CEB_1, \texttt{down}, CEB_2 \right)~| \\
            &~\texttt{ResEncoder}\left(E_{\text{Norm}} E_{\text{Nonlin}} E_{\text{Dropout}}, REB_1, \texttt{down}, REB_2 \right) \\
            3E~::=~&~\texttt{ConvEncoder}\left(E_{\text{Norm}} E_{\text{Nonlin}} E_{\text{Dropout}}, CEB_1, \texttt{down}, \dots, CEB_3 \right)~| \\
            &~\texttt{ResEncoder}\left(E_{\text{Norm}} E_{\text{Nonlin}} E_{\text{Dropout}}, REB_1, \texttt{down}, \dots, REB_3 \right) \\
            4E~::=~&~\texttt{ConvEncoder}\left(E_{\text{Norm}} E_{\text{Nonlin}} E_{\text{Dropout}}, CEB_1, \texttt{down}, \dots, CEB_4 \right)~| \\
            &~\texttt{ResEncoder}\left(E_{\text{Norm}} E_{\text{Nonlin}} E_{\text{Dropout}}, REB_1, \texttt{down}, \dots, REB_4 \right) \\
            2D~::=~&~\texttt{ConvDecoder}\left(D_{\text{Norm}} D_{\text{Nonlin}} D_{\text{Dropout}}, \texttt{up}, DB_1 \right) \\
            3D~::=~&~\texttt{ConvDecoder}\left(D_{\text{Norm}} D_{\text{Nonlin}} D_{\text{Dropout}}, \texttt{up}, DB_1, \texttt{up}, DB_2 \right) \\
            4D~::=~&~\texttt{ConvDecoder}\left(D_{\text{Norm}} D_{\text{Nonlin}} D_{\text{Dropout}}, \texttt{up}, DB_1, \dots, DB_3 \right) \\ 
            CEB_1~::=~&~\texttt{1b}~~|~~\texttt{2b}~~|~~\texttt{3b}~~|~~\texttt{4b} \\
            CEB_2~::=~&~\texttt{1b}~~|~~\texttt{2b}~~|~~\texttt{3b}~~|~~\texttt{4b} \\
            CEB_3~::=~&~\texttt{1b}~~|~~\texttt{2b}~~|~~\texttt{3b}~~|~~\texttt{4b} \\
            CEB_4~::=~&~\texttt{1b}~~|~~\texttt{2b}~~|~~\texttt{3b}~~|~~\texttt{4b} \\
            REB_1~::=~&~\texttt{1b}~~|~~\texttt{2b} \\
            REB_2~::=~&~\texttt{1b}~~|~~\texttt{2b}~~|~~\texttt{3b}~~|~~\texttt{4b}~~|~~\texttt{5b}~~|~~\texttt{6b} \\
            REB_3~::=~&~\texttt{1b}~~|~~\texttt{2b}~~|~~\texttt{3b}~~|~~\texttt{4b}~~|~~\texttt{5b}~~|~~\texttt{6b} \\
            ~&~\texttt{7b}~~|~~\texttt{8b} \\
            REB_4~::=~&~\texttt{1b}~~|~~\texttt{2b}~~|~~\texttt{3b}~~|~~\texttt{4b}~~|~~\texttt{5b}~~|~~\texttt{6b} \\
            ~&~\texttt{7b}~~|~~\texttt{8b}~~|~~\texttt{9b}~~|~~\texttt{10b}~~|~~\texttt{11b}~~|~~\texttt{12b} \\
            DB_1~::=~&~\texttt{1b}~~|~~\texttt{2b}~~|~~\texttt{3b}~~|~~\texttt{4b} \\
            DB_2~::=~&~\texttt{1b}~~|~~\texttt{2b}~~|~~\texttt{3b}~~|~~\texttt{4b} \\
            DB_3~::=~&~\texttt{1b}~~|~~\texttt{2b}~~|~~\texttt{3b}~~|~~\texttt{4b} \\
            DB_4~::=~&~\texttt{1b}~~|~~\texttt{2b}~~|~~\texttt{3b}~~|~~\texttt{4b} \\
            E_{\text{Norm}}~::=~&~\texttt{InstanceNorm}~~|~~\texttt{BatchNorm} \\
            E_{\text{Nonlin}}~::=~&~\texttt{LeakyReLU}~~|~~\texttt{ReLU}~~|~~\texttt{ELU}~~|~~\texttt{PReLU}~~|~~\texttt{GELU} \\
            E_{\text{Dropout}}~::=~&~\texttt{Dropout}~~|~~\texttt{NoDropout} \\
            D_{\text{Norm}}~::=~&~\texttt{InstanceNorm}~~|~~\texttt{BatchNorm} \\
            D_{\text{Nonlin}}~::=~&~\texttt{LeakyReLU}~~|~~\texttt{ReLU}~~|~~\texttt{ELU}~~|~~\texttt{PReLU}~~|~~\texttt{GELU} \\
            D_{\text{Dropout}}~::=~&~\texttt{Dropout}~~|~~\texttt{NoDropout} \\
        \end{aligned}$
    }
\end{equation}

\subsubsection{Prior-based Sampling of Architectures}
\label{sec:appendix:approach:hpo_hnas:prior_based_sampling}

In PriorBand~\citep{mallik-neurips23a}, prior-based sampling of hyperparameters is accomplished by drawing samples from a prior distribution.
The type and shape of the distribution are determined by
\begin{enumerate*}[label=(\roman*)]
    \item the type of hyperparameter and
    \item the prior confidence provided by the user.
\end{enumerate*}
For numerical hyperparameters, a truncated normal distribution is defined over the range of possible values, where the mean is set to the default value.
The prior confidence adjusts the standard deviation, with higher confidence resulting in a lower standard deviation.
Integer hyperparameters are sampled by rounding the values to the nearest integer value.
For categorical hyperparameters, in contrast, the probabilities are uniformly distributed across all values except for the default, whose probability is increased in proportion to the prior confidence.

\begin{figure}[tb]
	\includegraphics[width=\textwidth]{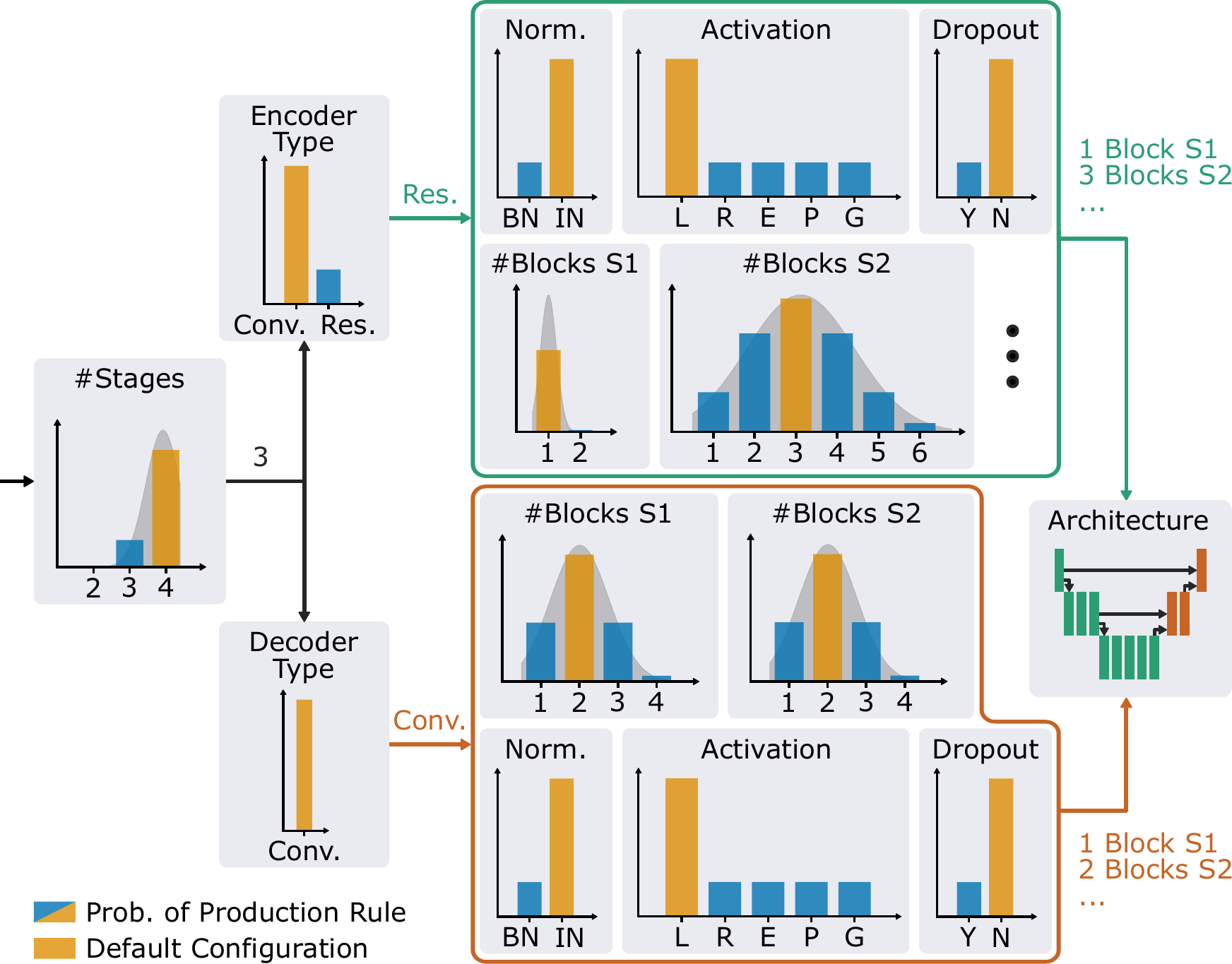}
	\caption{Overview of the prior-based sampling procedure for HNAS. Each block represents a design decision, i.e., the production rule of the CFG with its corresponding probability distribution. The probability of the possible value is indicated by the blue bar. The probability of sampling the default value is highlighted in orange. Notably, design decisions and default values differ based on previously selected values. Arrows indicate subsequent design decisions. First, the number of stages is sampled, then the encoder and decoder type are sampled. Based on the encoder type, the number of blocks per stage, as well as normalization, non-linearity, and dropout, are sampled. Abbreviations: BN = BatchNorm, IN = InstanceNorm, L = \texttt{LeakyReLU}, R = \texttt{ReLU}, E = \texttt{ELU}, P = \texttt{PreLU}, G = \texttt{GeLU}, Y = Yes (\texttt{True}), N = No (\texttt{False}).}
	\label{fig:approach:hnas:hnas_overview}
\end{figure}

To apply this concept to hierarchical architectures, we represent design decisions as integer and categorical hyperparameters.
For example, we model the type of encoder as a categorical hyperparameter with the convolutional encoder as the default value.
Based on the association of production rules with probability distributions proposed by \citet{schrodi-neurips23a}, we leverage the distributions of categorical and integer hyperparameters for the production rules.
Figure~\ref{fig:approach:hnas:hnas_overview} shows an overview of the prior-based sampling within the hierarchical NAS search space.
We consider an examplary search space with $n_{\text{stages,max}} = 4$ and $S_{\text{max}} = 2$.
We begin by sampling the number of stages using its associated production rule, which allows the U-Net to contain two, three, or four stages.
Since the default for this dataset is four, it is associated with the highest probability.
Here, we consider the network to consist of three stages.
Subsequently, the encoder and decoder are sampled.
In our example, we sample a residual encoder.
Thus, the subsequent distributions are computed based on the default block counts in \nnunet{} for a residual encoder.
Here, the first stage (S1) consists of a single block, whereas the second stage (S2) comprises three blocks.
For simplicity, we omit the number of blocks for the remaining stages in this example.
Similarly, the remaining design decisions are sampled for the decoder.

By following this approach, we are able to dynamically produce hierarchical prior distributions based on the corresponding default configurations in different branches within the search space.

\section{Experimental Setup}
\label{sec:appendix:experiments}

\subsection{Datasets}
\label{sec:appendix:datasets}
The Medical Segmentation Decathlon (MSD)~\citep{simpson-arxiv19,antonelli-nature22} is a collection of ten image segmentation datasets from the medical domain.
By focusing on diversity with respect to clinical tasks, modalities, and data characteristics, the MSD aims to serve as a standard for the evaluation of image segmentation algorithms.
The MSD is publicly available and provides access to all ten datasets for development and research purposes.
Live ranks of submissions are stated on the challenge leaderboard\footnote{\url{https://decathlon-10.grand-challenge.org/evaluation/challenge/leaderboard/}}.

\paragraph{Tasks, Modalities, and Characteristics}

\begin{figure}[htb]
	\includegraphics[width=\textwidth]{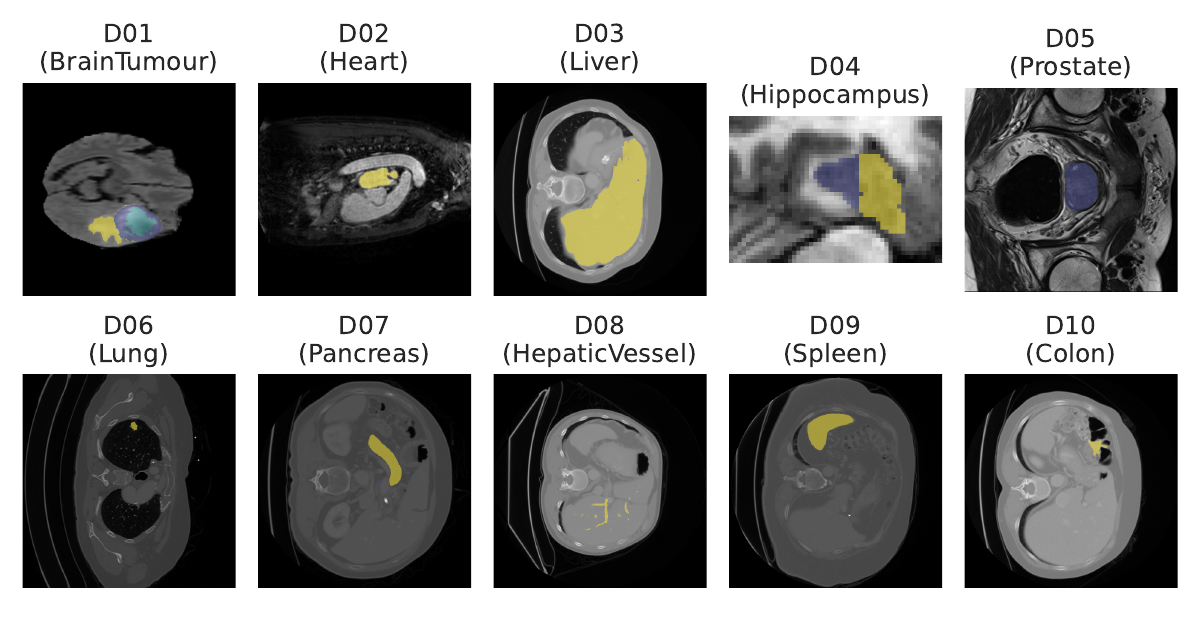}
	\caption{Example images from the MSD datasets with highlighted target labels, where each color represents an individual class. An image corresponds to the slice with the highest number of foreground voxels in the third image dimension. For 4D volumes, i.e., mp-MRI scans, the first parameter setting is selected.}
	\label{fig:experiments:datasets:msd_overview}
\end{figure}

Figure~\ref{fig:experiments:datasets:msd_overview} shows an overview of the ten datasets in the MSD, which we refer to as D01 to D10.
Each image contains a slice of a 3D volume with the target foreground labels highlighted.
For 4D volumes, the first modality is considered.
The MSD tasks cover a diverse range of segmentation tasks across different anatomical regions and imaging modalities.
Possible modalities are magnetic resonance imaging (MRI), computer tomography (CT), and multiparametric MRI (mp-MRI).
\Dds{01}\footnote{We use the original dataset name in British English even though this thesis is written in American English.}, for instance, shows brain tumor segmentations of an mp-MRI, while \Dds{03} and \Dds{06} contain segmentations of CT scans.
In addition, the size and structure of segmented areas vary between datasets.
\Dds{07}, we see fine-grained structures for two foreground target classes, whereas \Dds{02} features larger segmented areas corresponding to a single foreground target class.
This highlights the fundamental differences between datasets, which may necessitate specifically tailored models for each task to address their unique challenges and segmentation characteristics.

\begin{table}[htb]
    \centering
    \begin{tabular}{lrrrrrr}
        \toprule
		\textbf{Task} & \textbf{Name} & \textbf{Modality} & \textbf{\#Images} & \textbf{Shape / Dimensions} & \textbf{\#Classes} \\
        \midrule
		D01 & BrainTumour     & mp-MRI & 750 & [198,169,138] / 4D & 3 \\
		D02 & Heart           & MRI    & 30  & [115,320,232] / 3D & 1 \\
		D03 & Liver           & CT     & 201 & [432,512,512] / 3D & 2 \\
		D04 & Hippocampus     & MRI    & 394 & [36,50,35]    / 3D & 2 \\
		D05 & Prostate        & mp-MRI & 48  & [20,320,320]  / 4D & 2 \\
		D06 & Lung            & CT     & 96  & [252,512,512] / 3D & 1 \\
		D07 & Pancreas        & CT     & 420 & [93,512,512]  / 3D & 2 \\
		D08 & HepaticVessel   & CT     & 443 & [49,512,512]  / 3D & 2 \\
		D09 & Spleen          & CT     & 61  & [90,512,512]  / 3D & 1 \\
		D10 & Colon           & CT     & 190 & [95,512,512]  / 3D & 1 \\
        \bottomrule
	\end{tabular}
	\caption{MSD datasets with their respective characteristics. Shapes are median shapes after transposing the input images based on the dataset fingerprint of \nnunet{}~\citep{isensee-nature19}. For mp-MRI, the fourth dimension contains the sequence of MRI scans using different parameters. The number of classes refers to the number of foreground labels.}
	\label{tab:appendix:related:msd_datasets_short}
\end{table}

Table~\ref{tab:appendix:related:msd_datasets_short} states the metadata of the ten tasks in the MSD, highlighting their key characteristics.
An mp-MRI scan contains a sequence of MRI scans captured with different parameter settings, introducing an additional image dimension.
The datasets also vary in size, resolution, and number of segmentation classes, leading to diverse challenges for evaluating MIS methods.

\paragraph{Evaluation Protocol}
Each task in the MSD is divided into a fixed training and test set.
Only input images are provided for the test set, and the corresponding labels are unavailable. 
After fitting a model on the training set, participants need to generate predictions for the test set and upload them to an online evaluation platform\footnote{\url{https://decathlon-10.grand-challenge.org}}.
Since test set labels are not publicly available, the platform is the only method for evaluating a model on the test set.
It then returns the test set accuracy using the Dice Similarity Coefficient (DSC)~\citep{dice_ecology45}.

\subsection{AutoML Methods}
\label{sec:appendix:experiments:automl_methods}

\subsubsection{Additional Baseline}
Recent work on foundation models for computer vision has led to their application in the medical domain.
\textit{Segment Anything}~(SAM)~\citep{kirillov-iccv23} is an image segmentation foundation model pre-trained on a dataset containing 1M images and 1B ground-truth segmentation masks.
Unlike task-specific models, e.g., U-Nets, which need to be trained from scratch for each new segmentation dataset, the pre-training enables foundation models to generalize across diverse datasets and reduces the need for extensive labeled medical data.

\medsam{}~\citep{ma-nature24}, based on SAM, fine-tuned on large-scale medical imaging data, is a foundation model for MIS and can outperform task-specific models.
However, as 3D image segmentations must be obtained by segmenting individual 2D slices, \medsam{} achieves limited accuracy for 3D images.
To overcome this limitation, \medsamtwo{}~\citep{ma-arxiv24} facilitates a transfer-learning pipeline for SAM2~\citep{ravi-arxiv24}.
SAM2 is a recent foundation model built on SAM for promptable image and video segmentation, trained on 35.5M masks from 50.9K videos.
It replaces the vision transformer (ViT)~\citep{dosovitskiy-iclr21} in SAM with a hierarchical ViT~\citep{ryali-icml23} and adds a memory attention module to condition the current frame on the previous one.
The video segmentation capabilities of SAM2 enable \medsamtwo{} to represent 3D volumes as a sequence of 2D frames and produce improved 3D medical image segmentations compared to \medsam{}.

SAM2 was originally designed for interactive image and video segmentation and requires a prompt, i.e., a user input, to identify the object to segment.
During training, the \medsamtwo{} framework facilitates box prompts, which add bounding boxes around target segmentations alongside each input frame, i.e., a slice of a 3D image.
During inference, the frame with the largest bounding box enclosing the target segmentation area is selected as the starting frame.
The model is then prompted with two sequences of frames:
one spanning from the starting frame to the first frame of the volume and the other extending from the starting frame to the last frame.
The separate predictions from these sequences are aggregated to obtain the final segmentation mask. 
Notably, to determine the starting frame, this inference method relies on ground truth segmentations that require additional effort to obtain.

We leverage the pipeline proposed by the authors to finetune MedSAM2 on each individual MSD dataset.
Due to resource limitations, we reduce the number of training epochs from $1\,000$ to $100$, which is roughly equivalent to the training runtime of the most expensive \nnunet{} configuration on \Dds{01}, the dataset with the highest training runtime.
Furthermore, we incorporate intermediate model evaluations on the validation split as accuracy estimates throughout the training process.

\subsubsection{PriorBand Setup}
\label{sec:appendix:experiments:automl_methods:priorband}

For PriorBand, we rely on the setup proposed by \citet{mallik-neurips23a} with the number of training epochs as HB budget.
Given that the number of epochs is set to $1\,000$ by default in nnU-Net, we set $b_{\text{min}} = 10$ and $b_{\text{max}} = 1\,000$.
We set the reduction factor $\eta$ to the default value of $3$ as proposed by \citet{li-iclr17a} and \citet{mallik-neurips23a}.
We round budgets to full epochs.
With the initial evaluation of the default configuration at the maximum fidelity, this leads to $129$ evaluated configurations and a total budget of $22\,000$ epochs.
As we continue runs within SH to reduce the computational demands, this results in a total of $18\,308$ trained epochs for an optimization run, excluding 5-fold cross-validation.

\subsubsection{Search Spaces}
\label{sec:appendix:experiments:automl_methods:search_spaces}

In the following, we state details on hyperparameters.
\textit{Optimizer} can be \textit{stochastic gradient descent with momentum}~(\texttt{SGD})~\citep{goodfellow-16a}, \texttt{Adam}~\citep{kingma-iclr15a}, or \texttt{AdamW}~\citep{loshchilov-iclr19a}. \textit{Momentum} is only enabled for \texttt{SGD}. \textit{Learning Rate Scheduler} can use a polynomial schedule~(\texttt{PolyLRScheduler})~\citep{mishra-ieee19}, cosine annealing schedule~\citep{loshchilov-iclr17a}, or no schedule at all~(\texttt{None}). \textit{Foreground Oversampling} defines the proportion of samples in each batch that must contain foreground segmentations. \textit{Data Augmentation Factor} sets a multiplier that is applied to each individual data augmentation probability. When set to $0$, no data augmentation is applied.
\textit{Model Scale} defines the scale by multiplying the default number of blocks per stage in the U-Net. Notably, ordinal hyperparameters are modeled as integer values mapped to actual hyperparameter values. For the encoder, the default changes based on the encoder type. \textit{Base \#Features} defines the number of features on base, i.e., the input and output stage of the U-Net. \textit{Max. \#Features} defines the maximum number of features in the bottleneck of the U-Net. When constructing the network, the number of features is doubled for each subsequent stage, but the maximum number is an upper bound. \textit{Activation} can be \textit{rectified linear unit}~(\texttt{ReLU})~\citep{nair-icml10a}, \texttt{LeakyReLU}~\citep{maas-icml13}, \textit{exponential linear unit}~(\texttt{ELU})~\citep{clevert-iclr16}, \textit{gaussian error linear unit}~(\texttt{GELU})~\citep{hendrycks-arxiv16}, and \textit{parametric ReLU}~(\texttt{PReLU})~\citep{he-iccv15}. \textit{Normalization} can be \textit{batch normalization} (\texttt{BatchNorm})~\citep{ioffe-icml15a} or \textit{instance normalization} (\texttt{InstanceNorm})~\citep{ulyanov-arxiv16}.

\begin{table}[h]
    \centering
	\begin{tabular}{l|lrrr}
        \toprule
        \textbf{Type} & \textbf{Hyperparameter} & \textbf{Type} & \textbf{Range / Values} & \textbf{Default Value} \\
        \midrule
        HPO & Optimizer & Categorical & \{\texttt{SGD}, \texttt{Adam}, \texttt{AdamW}\} & \texttt{SGD} \\
        \midrule
        HPO & Momentum (SGD) & Float (log) & [$0.5$, $0.999$] & $0.99$ \\
        \midrule
        HPO & \makecell[l]{Initial\\Learning Rate} & Float (log) & [$1\cdot10^{-5}$, $0.1$] & $1\cdot10^{-2}$ \\
        \midrule
        HPO & \makecell[l]{Learning Rate\\Scheduler} & Categorical & \makecell[r]{\{\texttt{PolyLRScheduler},\\ 
        \texttt{CosineAnnealingLR},\\ \texttt{None}\}} & \texttt{PolyLRScheduler} \\
        \midrule
        HPO & Weight Decay & Float (log) & [$1\cdot10^{-6}$, $1\cdot10^{-2}$] & $3\cdot10^{-5}$ \\
        \midrule
        HPO & \makecell[l]{Foreground\\Oversampling} & Float & [$0$, $1$] & $0.33$ \\
        \midrule
        HPO & Loss Function & Categorical & \makecell[r]{\{\texttt{DiceLoss},\\ \texttt{CrossEntropyLoss},\\
        \texttt{DiceAndCross}-\\ \texttt{EntropyLoss},\\  \texttt{TopKLoss}\}} & \makecell[r]{\texttt{DiceAndCross}-\\ \texttt{EntropyLoss}} \\
        \midrule
        HPO & \makecell[l]{Data Augmentation \\ Factor} & Float & [$0$, $3$] & $1$ \\
        \midrule
        \midrule
        NAS & Encoder Type & Categorical & \makecell[r]{\{\texttt{Convolutional}-\\\texttt{Encoder},\\ \texttt{ResidualEncoderM}\}} & \makecell[r]{\texttt{Convolutional}-\\\texttt{Encoder}} \\
        \midrule
        NAS & Model Scale & Ordinal & [$0.5$, $1$, $1.5$, $2$] & $1$ \\
        \midrule
        NAS & \makecell[l]{Base \#Features} & Integer & [$16$, $64$] & $32$ \\
        \midrule
        NAS & \makecell[l]{Max. \#Features} & Integer & [$160$, $640$] & $320$ \\
        \midrule
        NAS & Activation & Categorical & \makecell[r]{\{\texttt{LeakyReLU}, \texttt{ReLU},\\ \texttt{ELU}, \texttt{GELU}, \texttt{PReLU}\}} NAS & \texttt{LeakyReLU} \\
        \midrule
        NAS & Normalization & Categorical & \makecell[r]{\{\texttt{BatchNorm},\\ \texttt{InstanceNorm}\}} & \texttt{InstanceNorm} \\
        \midrule
        NAS & Dropout Rate & Float & [$0$, $0.5$] & $0$ \\
        \bottomrule
        \bottomrule
	\end{tabular}
	\caption{HPO \textbf{(top)} and NAS \textbf{(bottom)} hyperparameters in the JAHS search space in \autonnunet{}.}
	\label{tab:experiments:setup:jahs_search_space}
\end{table}

\begin{table}[H]
    \centering
	\begin{tabular}{lrrr}
        \toprule
        \textbf{Hyperparameter} & \textbf{Type} & \textbf{Range / Values} & \textbf{Default Value} \\
        \toprule
        Dropout Rate & Float & [$0$, $0.5$] & $0.2$ \\
        \midrule
        Architecture & CFG-Architecture & - & - \\
        \bottomrule
	\end{tabular}
	\caption{Additional HNAS hyperparameters in the \hpohnas{} search space, replacing the NAS hyperparameters in the \hpohnas{} search space. The context-free grammar-based architecture~(\textit{CFG-Architecture}, \cite{schrodi-neurips23a}) defines the neural architecture using function compositions (see Section~\ref{sec:approach:hpo_hnas}).}
	\label{tab:experiments:setup:hpo_hnas_search_space}
\end{table}

\subsection{Experimental Pipeline}
\label{sec:appendix:experiments:pipeline}

Our \textit{AutoNNUNet} package builds the entry point for all experiments and visualizations.
For the baseline models, we rely on adaptions of the \textit{nnunetv2}~\citep{isensee-nature19}, \textit{MedSAM}~\citep{ma-nature24}, and \textit{batchgenerators}~\citep{isensee_github20} packages.
These adaptions add support for running the frameworks on compute clusters.
PriorBand and regularized PriorBand are implemented in our extension of the \textit{Neural Pipeline Search}~(NePS)~\citep{stol_neps23} framework.
Our adaption of the \textit{HyperSweeper}~\citep{eimer-github24} framework integrates multi-objective optimization methods.
All models are trained and evaluated using 5-fold cross-validation based on the splits obtained by \nnunet{} during its planning phase.
Thus, we use the exact same splits for all baseline and optimization experiments.

\section{Additional Results}
\label{sec:appendix:results}

\begin{table}[h]
    \centering
    \begin{tabular}{l|lll|l|l|ll}
        \toprule
        Approach & \multicolumn{3}{c|}{\textbf{\nnunet{}}} & \multicolumn{1}{c|}{\textbf{\medsamtwo{}}} & \makecell{\textbf{Auto-}\\ \textbf{nnU-Net}} & \multicolumn{2}{c}{\textbf{Ablations}} \\
        \cmidrule(r@{2pt}){2-4}                        
        \cmidrule(l@{2pt}){7-8}
         & Conv & ResM & ResL & & & HPO & \makecell{HPO +\\HNAS} \\
        Dataset &  &  &  &  &  &  &  \\
        \midrule
        D01 (BrainTumour) & $73.98$ & $74.15$ & $73.60$ & $43.87$ & $\mathbf{74.45}$ & $74.21$ & $74.35$ \\
        D02 (Heart) & $93.39$ & $93.40$ & $93.26$ & $87.66$ & $\mathbf{93.53}$ & $93.43$ & $93.39$ \\
        D03 (Liver) & $79.45$ & $\mathbf{81.66}$ & $81.59$ & $65.26$ & $80.36$ & $79.58$ & $79.45$ \\
        D04 (Hippocampus) & $89.04$ & $88.75$ & $88.62$ & $69.52$ & $\mathbf{89.46}$ & $89.37$ & $89.25$ \\
        D05 (Prostate) & $73.53$ & $73.64$ & $72.97$ & $62.21$ & $75.23$ & $\mathbf{75.30}$ & $74.87$ \\
        D06 (Lung) & $68.33$ & $68.03$ & $68.58$ & $68.32$ & $69.19$ & $\mathbf{71.01}$ & $69.73$ \\
        D07 (Pancreas) & $66.07$ & $67.78$ & $67.82$ & $61.82$ & $67.05$ & $67.13$ & $\mathbf{67.83}$ \\
        D08 (HepaticVessel) & $68.31$ & $\mathbf{68.66}$ & $67.67$ & $45.39$ & $68.31$ & $68.31$ & $68.31$ \\
        D09 (Spleen) & $96.66$ & $96.76$ & $\mathbf{97.03}$ & $93.87$ & $96.76$ & $97.02$ & $96.92$ \\
        D10 (Colon) & $46.04$ & $44.05$ & $50.47$ & $\mathbf{78.96}$ & $51.98$ & $46.03$ & $46.03$ \\
                \midrule
        \textbf{Mean} & $75.48$ & $75.69$ & $76.16$ & $67.69$ & $\mathbf{76.63}$ & $76.14$ & $76.01$ \\
        \bottomrule
    \end{tabular}
    \caption{Mean 5-fold cross-validation DSC [\%] based on the nnU-Net dataset splits for baseline and AutoML incumbent configurations. The best-performing method per dataset is highlighted in \textbf{bold}.}
    \label{tab:results:dsc_overview}
\end{table}

\begin{figure}[h]
    \centering
    \includegraphics[width=0.49\textwidth]{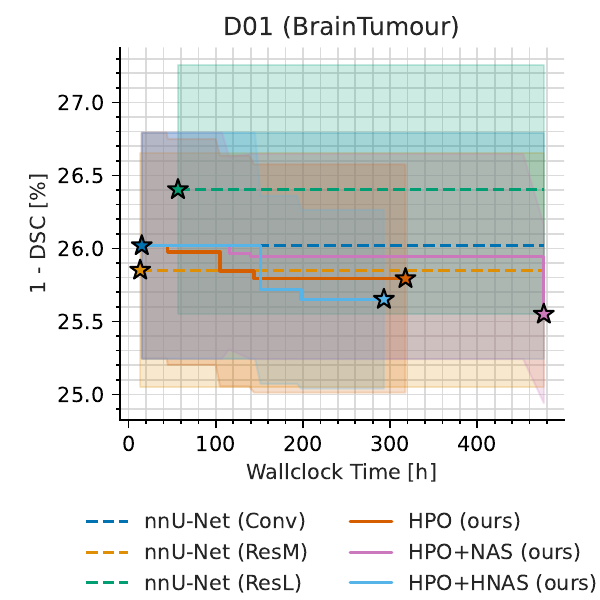}
    \includegraphics[width=0.49\textwidth]{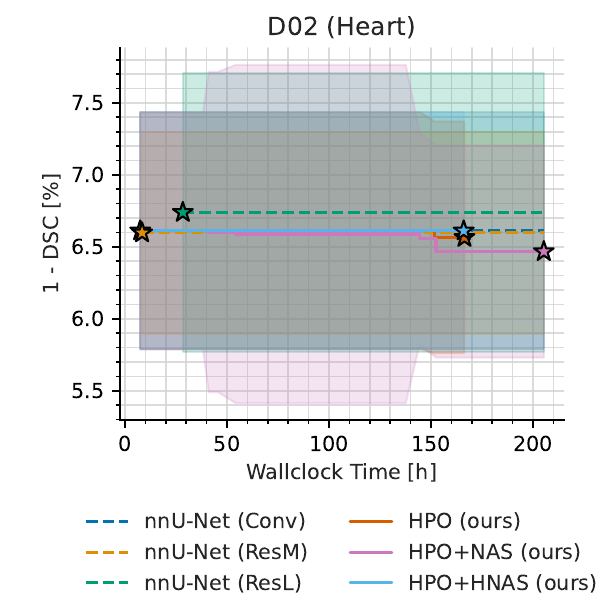}
    \includegraphics[width=0.49\textwidth]{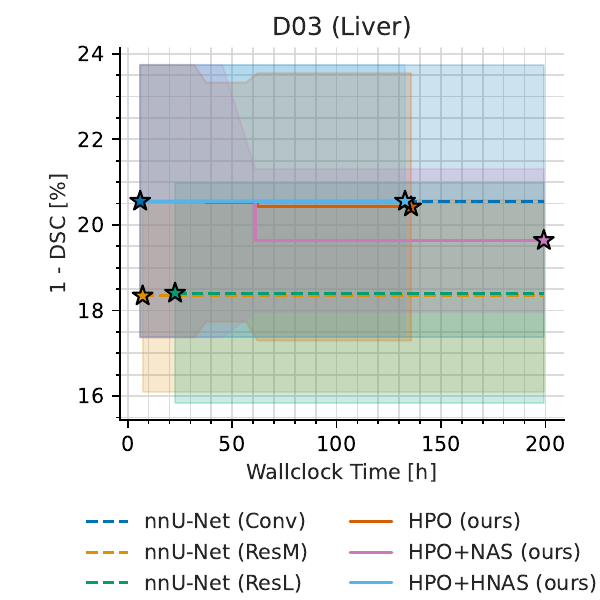}
    \includegraphics[width=0.49\textwidth]{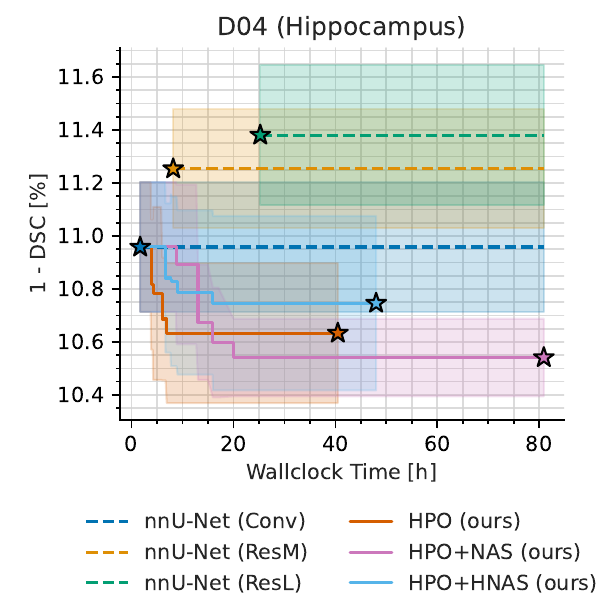}
    \caption{Optimization performance over time. Error bars indicate standard deviation across 5-fold cross-validation splits.}
    \label{fig:appendix:results:performance_over_time_1}
\end{figure}

\begin{figure}[h]
    \centering
    \includegraphics[width=0.49\textwidth]{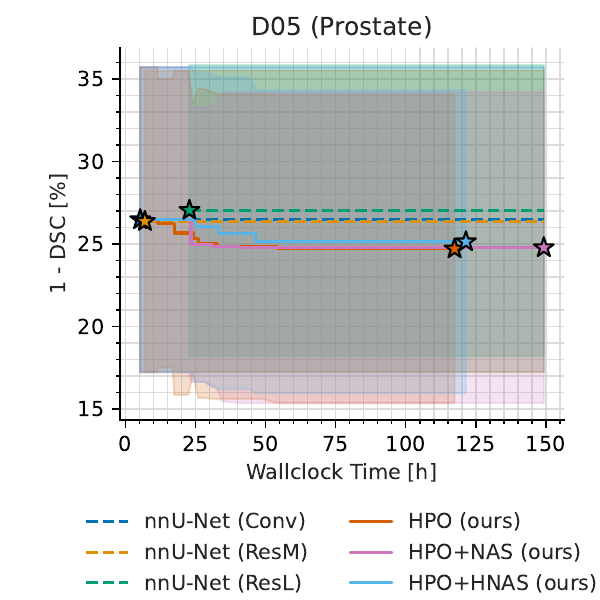}
    \includegraphics[width=0.49\textwidth]{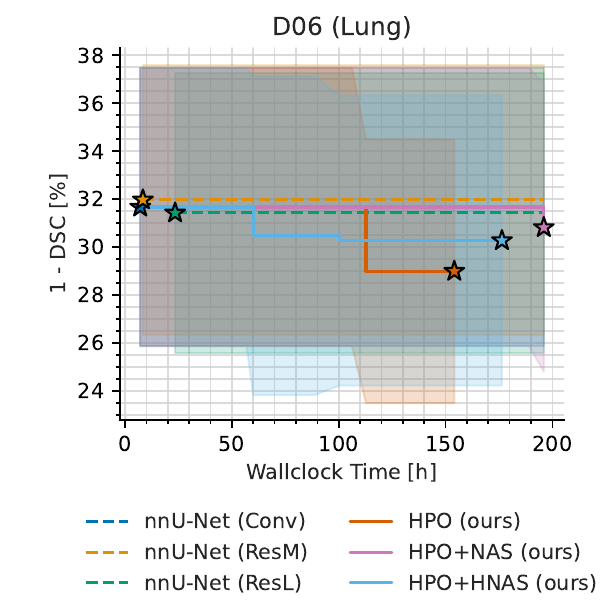}
    \includegraphics[width=0.49\textwidth]{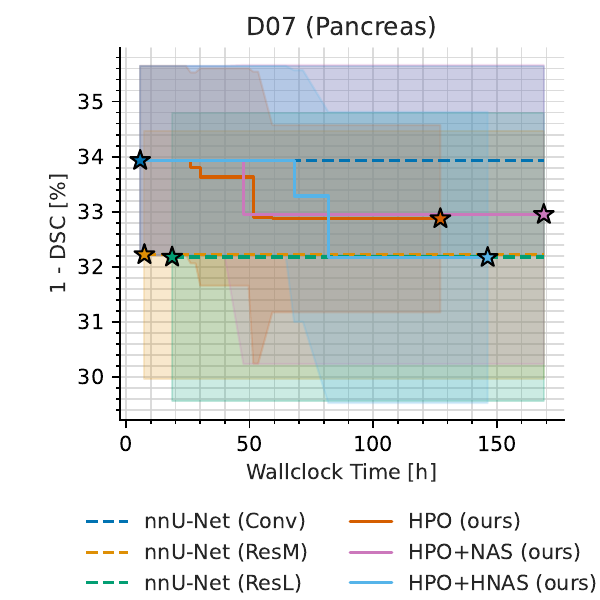}
    \includegraphics[width=0.49\textwidth]{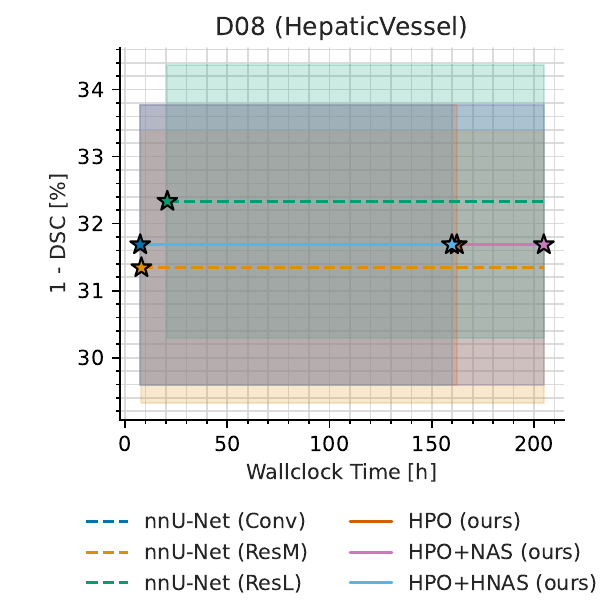}
    \caption{Optimization performance over time. Error bars indicate standard deviation across 5-fold cross-validation splits.}
    \label{fig:appendix:results:performance_over_time_2}
\end{figure}

\begin{figure}[h]
    \centering
    \includegraphics[width=0.49\textwidth]{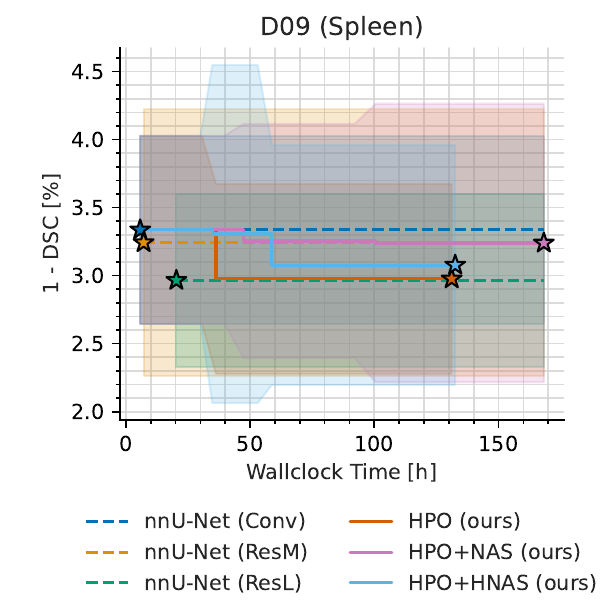}
    \includegraphics[width=0.49\textwidth]{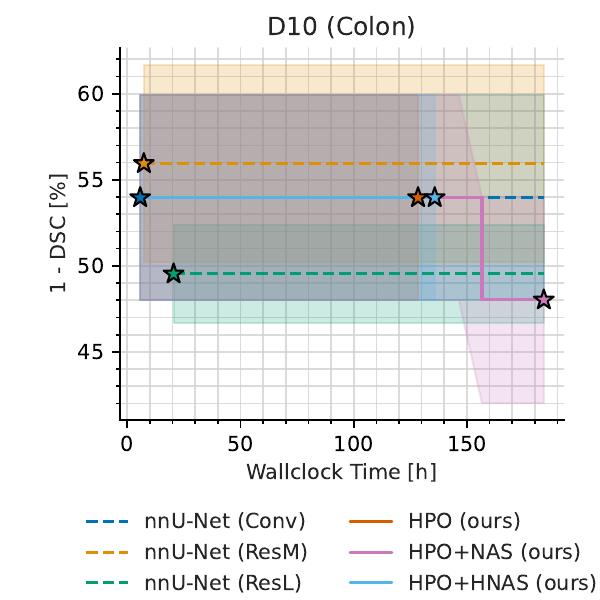}
    \caption{Optimization performance over time. Error bars indicate standard deviation across 5-fold cross-validation splits.}
    \label{fig:appendix:results:performance_over_time_3}
\end{figure}

\begin{figure}[h]
    \centering
    \includegraphics[width=\linewidth]{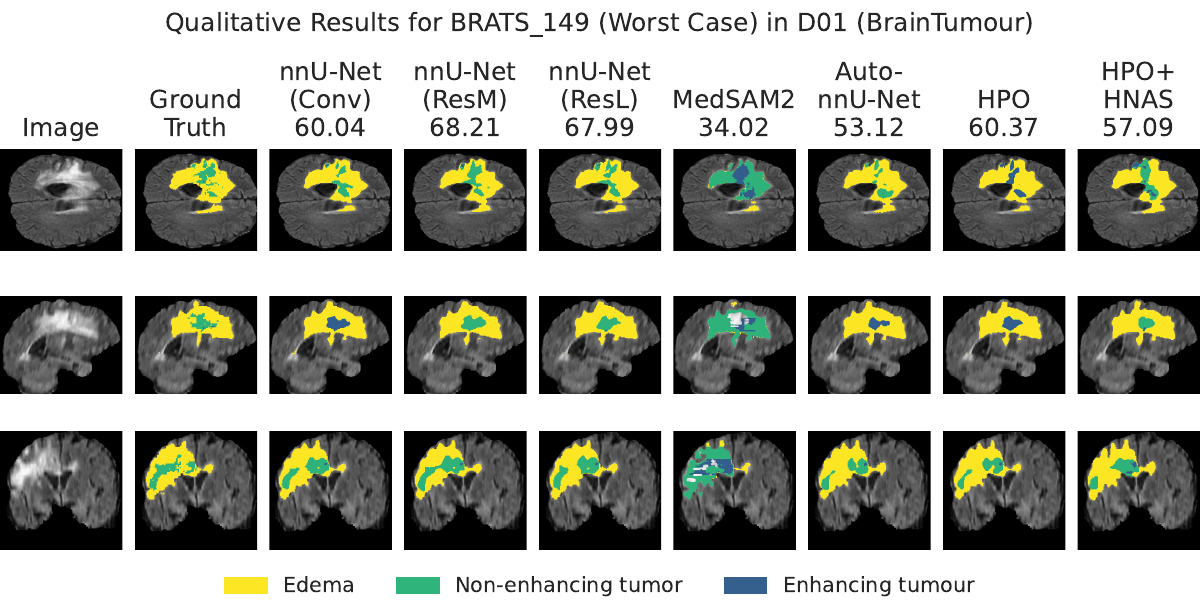}
    \caption{Qualitative segmentation results for \Dds{01}. The columns correspond to the input image, ground truth segmentation mask, and predicted segmentations of the methods, where colors represent foreground classes. Numbers below the method names correspond to their respective DSC in \% for this example. Each row of the figure represents a slice of the 3D volume along one axis. As this 4D volume is an mp-MRI scan, the first parameter setting is selected, yielding a 3D volume. Results for all datasets are stated in our GitHub repository.}
    \label{fig:results:automl_methods:overview:qualitative_d01_worst}
\end{figure}

\begin{figure}[h]
    \centering
    \includegraphics[width=0.49\textwidth]{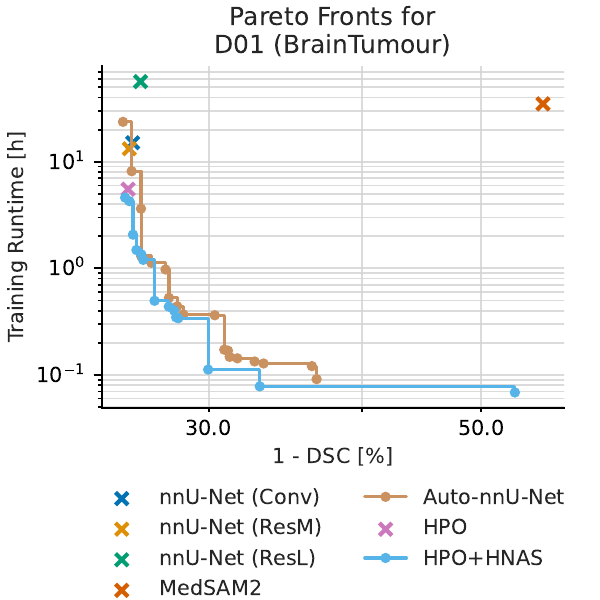}
    \includegraphics[width=0.49\textwidth]{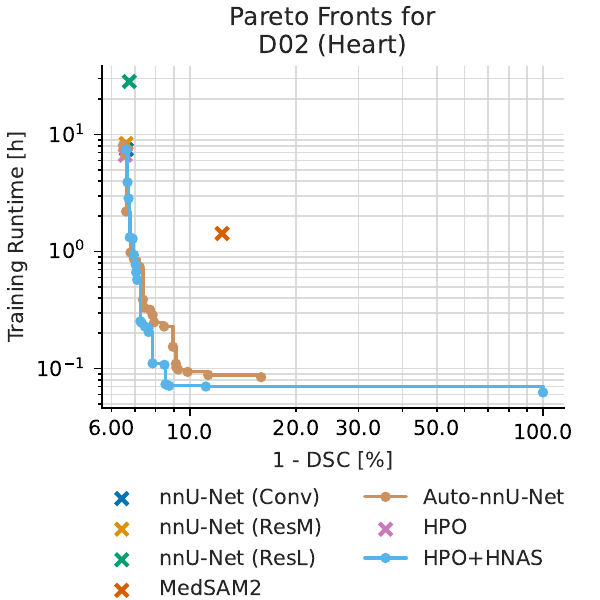}
    \includegraphics[width=0.49\textwidth]{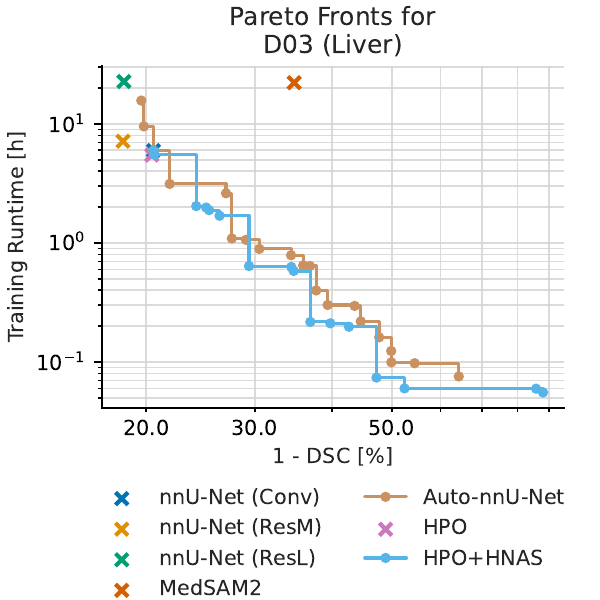}
    \includegraphics[width=0.49\textwidth]{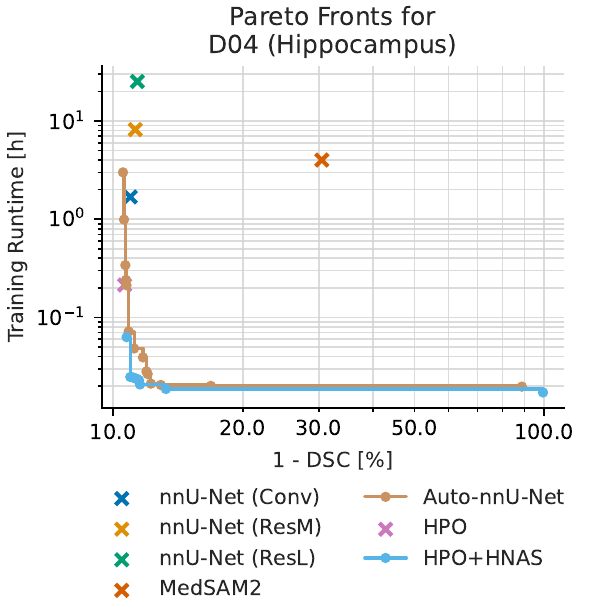}
    \includegraphics[width=0.49\textwidth]{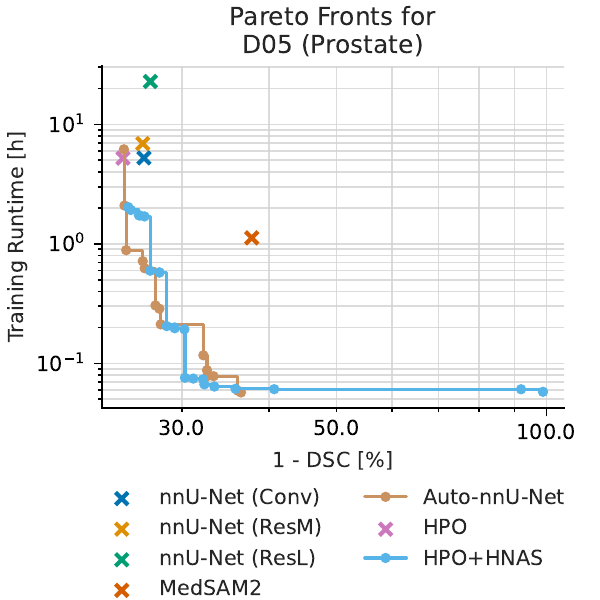}
    \includegraphics[width=0.49\textwidth]{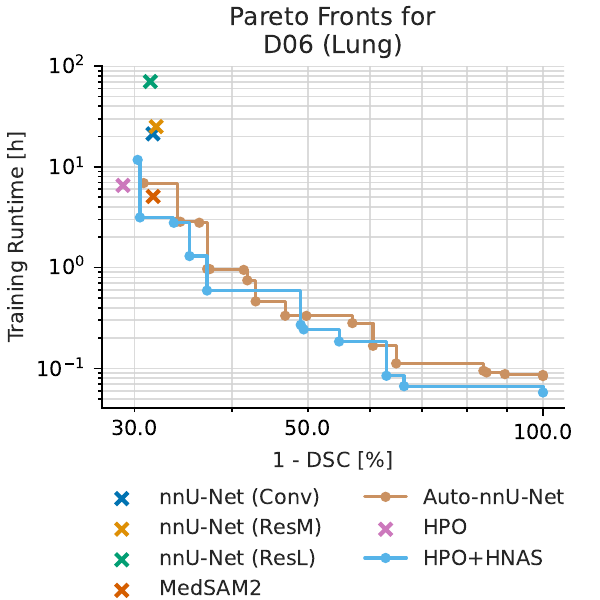}
    \caption{Pareto fronts of \hponas{} and \hpohnas{}  compared to the baselines and HPO results.}
    \label{fig:appendix:results:pareto_fronts_1}
\end{figure}

\begin{figure}[h]
    \centering
    \includegraphics[width=0.49\textwidth]{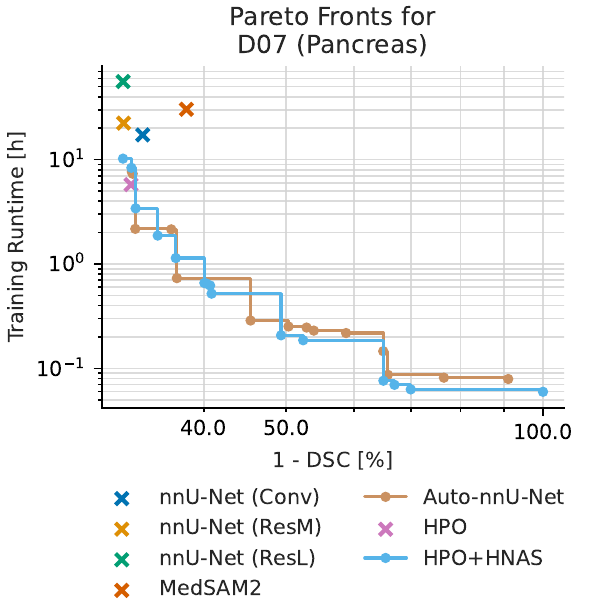}
    \includegraphics[width=0.49\textwidth]{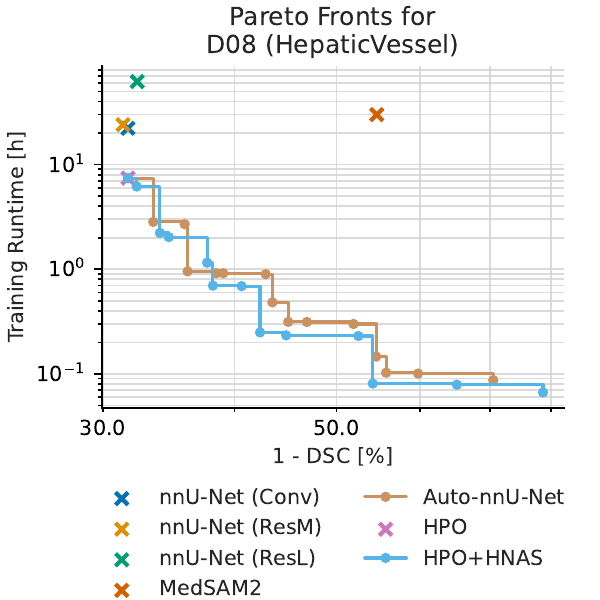}
    \includegraphics[width=0.49\textwidth]{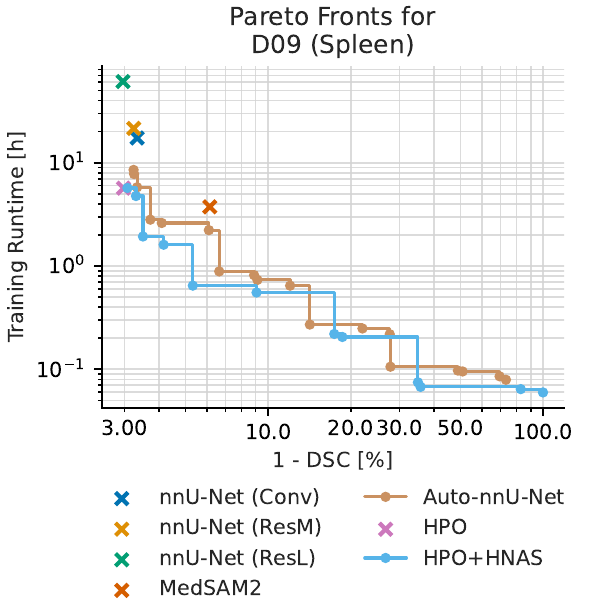}
    \includegraphics[width=0.49\textwidth]{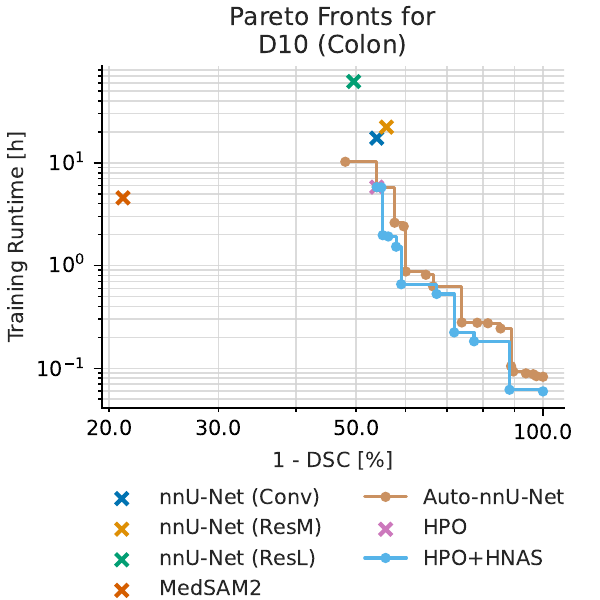}
    \caption{Pareto fronts of \hponas{} and \hpohnas{}  compared to the baselines and HPO results.}
    \label{fig:appendix:results:pareto_fronts_2}
\end{figure}

\begin{figure}[h]
    \centering
    \includegraphics[width=\linewidth]{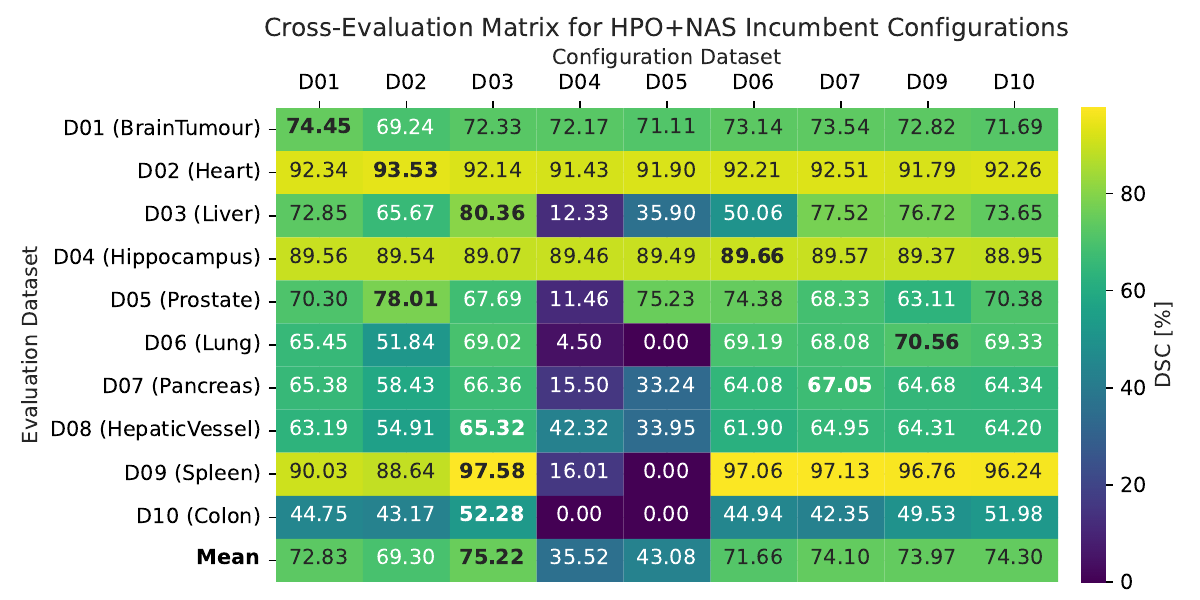}
    \caption{Cross-evaluation matrix for \hponas{} incumbent configurations. Each cell states the 5-fold cross-validation DSC [\%] when applying an incumbent configuration of a dataset (column) to a different dataset (row). In addition, the mean per incumbent configuration is stated. The highest accuracy per evaluation dataset is indicated in \textbf{bold}.}
    \label{fig:appendix:results:automl_results:cross_eval_matrix}
\end{figure}

\begin{figure}[h]
    \centering
    \includegraphics[width=\linewidth]{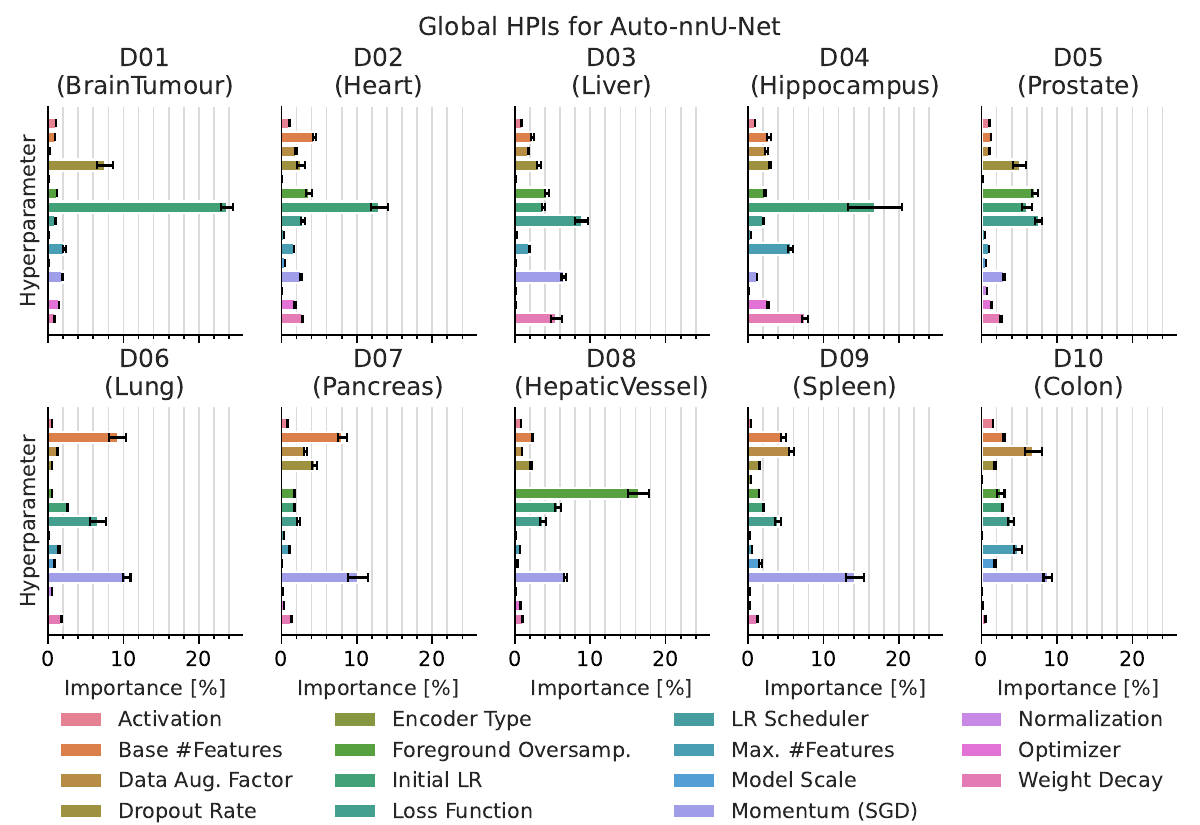}
    \caption{Global functional ANOVA~(fANOVA)~\citep{hutter-icml14a} hyperparameter importance \autonnunet{} across all datasets for \odsc{} with error bars indicating variances.}
    \label{fig:results:automl_analysis:hpo:global_hpis}
\end{figure}

\end{document}